\UseRawInputEncoding
\documentclass[11pt,3p,review]{elsarticle}
\makeatletter
\def\ps@pprintTitle{%
 \let\@oddhead\@empty
 \let\@evenhead\@empty
 \def\@oddfoot{\centerline{\thepage}}%
 \let\@evenfoot\@oddfoot}
\makeatother
\newgeometry{left=1.5in,right=1.5in,top=1.5in,bottom=1.5in}

\biboptions{longnamesfirst}

\usepackage[numbers]{natbib}
\usepackage{epsfig}
\usepackage{amsfonts}
\usepackage{booktabs}
\usepackage{subcaption}
\usepackage{xcolor}
\usepackage{amssymb}
\usepackage{graphicx}
\graphicspath{{images/}}

\newtheorem {assumption}{Assumption}
\newtheorem{definition}{Definition}
\newtheorem{theorem}{Theorem}
\newtheorem{remark}{Remark}
\newtheorem{lemma}{Lemma}

\begin{document}
\begin{frontmatter}

\title{Short-term prediction of Time Series based on bounding techniques}

\author[ss]{Pedro Cadah\'ia \corref{cor}}
\ead{pedro.cadahia@alu.uhu.es}
\author[ss]{Jos\'e M. Bravo}

\address[ss]{Escuela T\'ecnica Superior de Ingenier\'ia, Universidad de Huelva,\\
Carretera Huelva - Palos de la Frontera s/n. 21819. La R\'abida - Palos de la Frontera. Huelva. Spain}

\cortext[cor]{Correspondence: Pedro Cadah\'ia, Escuela T\'ecnica Superior de Ingenier\'ia, Universidad de Huelva,\\
Carretera Huelva-Palos de La Frontera s/n. 21819. Huelva, Spain.}

\begin{abstract}
In this paper it is reconsidered the prediction problem in time series framework by using a new non-parametric approach. Through this reconsideration, the prediction is obtained by a weighted sum of past observed data. These weights are obtained by solving a constrained linear optimization problem that minimizes an outer bound of the prediction error. The innovation is to consider both deterministic and stochastic assumptions in order to obtain the upper bound of the prediction error, a tuning parameter is used to balance these deterministic-stochastic assumptions in order to improve the predictor performance. A benchmark is included to illustrate that the proposed predictor can obtain suitable results in a prediction scheme, and can be an interesting alternative method to the classical non-parametric methods. Besides, it is shown how this model can outperform the preexisting ones in a short term forecast.
\end{abstract}


\end{frontmatter}

\section{Introduction}
The purpose of this paper is to provide a new model for time series set up on the observed past values of the time series, by means of a non-parametric approach. It is well-known fact that in parametric time series analysis the relationship between observed past values of the time series and the prediction is defined by specifying a functional form and a fixed finite number of parameters. Widely studied parametric options are auto-regressive (AR) models, moving average (MA) models, and different combinations as ARMA or ARIMA models \cite{BoxJen76,Hamilton94}. In nonlinear time series, some common parametric structures has been studied, the threshold auto-regressive (TAR) models \cite{Tong83}, the exponential auto-regressive (EXPAR) model and smooth-transition auto-regressive (STAR) models  are some examples \cite{Haggan81, Chang86}. The performance of the parametric predictor is a consequence of the a priori function form chosen.

By contrast, in non-parametric approaches a more flexible class of functions is considered. Non-parametric methods avoid the choosing of a specific functional form. Collected data provides the information to obtain a new prediction. The price to pay is the 'curse of dimensionality', that is, a possible poor performance in high dimensions prediction problems. Local conditional mean or median method provides a prediction using the mean or the median of a neighborhood of the interest point \cite{Truong93}. The Nadaraya-Watson estimator averages past observations by a kernel function to obtain a prediction \cite{Narayada64,Wats:1964}. Local linear o polynomial functions of past observations can be used to approximate a nonlinear relationships \cite{Hardle90,FanGij96}. Semi-parametric models as nonlinear additive auto-regressive (NAAR) models or functional coefficient auto-regressive (FAR) models have been proposed too \cite{hastib90, Hardle97}. Many researchers have written an extensive review of non-parametric methods applied to time series prediction \cite{FanYao03,Gao07,GOOIJER2000259, YIN2016266}.

In this paper a new non-parametric prediction method is proposed. The prediction is obtained by a weighted sum of past observations. An upper bound of the prediction error is computed under some deterministic and stochastic assumptions. A constrained optimization problem is formulated to minimize the upper bound of the prediction error and to obtain the set of optimal weights used to compute the prediction. The optimization problem includes a parameter to balance the deterministic-stochastic assumptions. This is the main novelty of the proposed method. This parameter can be tuned with training data and a cross-validation scheme to improve the predictor performance \cite{BERGMEIR201870}. The proposed predictor provides a general framework that encompasses some relevant non-parametrics predictors as the Nadaraya-Watson predictor \cite{Narayada64,Wats:1964} or predictors based on local linear regression \cite{FanGij96}, these models have been widle used in the literature \cite{MANGALOVA20161023, Hyndman2004LocalLF}.

The paper is organized as follows. In Section \ref{SecForm}, the problem formulation is addressed. The deterministic and stochastic assumptions are  presented in Section \ref{S3}. The new predictor is proposed in Section \ref{secVI}. Benchmark results are illustrated in Section \ref{ejemplo}. Finally, Section \ref{conclusion} reports some conclusions.

\section{Formulation}\label{SecForm}
It is considered a discrete\footnote{It is assumed a discrete version of data.} time series process $\{Z_t\}$ with $t\in \{0,\pm1,\pm2,\ldots\}$. At time instant $k$ it is assumed that past data $\{Z_t\}$ with $t \in \{k,k-1,k-2,...\}$ has been observed and there is interest in providing a forecast for predicting $Z_{k+1}$. Once the detrend is applied to the time series,\footnote{It should be noted that in coherence with the prediction system and in order to estimate $\mu_{k+1}$, only the past observations can be used, independently of the detrending method used.} the time series is now the series $\{y_t\}$ with $t \in \{k,k-1,...\}$, where $Z_t = y_t + \mu_t$, being $\mu_t$ the trend component and $y_{k+1}$ the detrended future time series value.

It is also denoted by $\{z_j\}$ with  ${j=0,1,...,k}$ the set of the vectors consisting of the observed past values of the time series, that is $z_j=[y_{j},y_{j-1},\ldots , y_{j-p-1}]^{T}$ . Henceforth this $p$-dimensional vector set will be called {\it embedding vector}. This set of data is used to forecast future values for the time series. It is a must to clarify this point in order to precise the sense of the parametric and non-parametric models used in this article. A parametric approach is characterized by the use of the training set for estimating the parameters of the model and once this inference is done the data set is not used again. The non-parametric approach considered in this work, it is a local approach in which each forecast is obtained by using all the available data set but selecting a neighborhood of the interest point. In this sense, it is assumed that the time series can be generated by an unknown local linear model.

\begin{assumption} \label{ASS_1}
Considering the forecast of $y$ modeled as:
\begin{equation}
	y_{k+1} = r(z_k)^T \Phi_k + e_k                 \label{ass1eq1}
\end{equation}
where it is assumed that the existence of an unknown vector of parameters $\Phi_k\in \mathcal{R}^n$, a known function $r(\cdot)$ valuated at the {\it embedding set} and an unknown error term $e_k$.\footnote{This modeling is flexible enough to admit alternative assumptions about the error term. As discussed later, the model is presented by using both deterministic and stochastic bounds for the error term $e_k$.}
\end{assumption}

In order to complete the presentation of the model it should be discussed in more detail the so called {\it regressor generator function} $r(\cdot)$. This function allows transform the original values into vectors of dimension $n_r$ by means of the vectors belonging to the embedding set. A formal definition of this regressor generator function is as follows.

\begin{definition}[Regressor generator function]\label{def1}
The function $r(\cdot):\mathcal{R}^{p} \rightarrow \mathcal{R}^{n_r} $ specifies the regressor vector components. This function admits any kind of auto-regressive representation, nonlinear expression of past components and different functional forms for decomposing the different components of the time series.\footnote{For instance suppose a set $z_k=[y_{k},y_{k-1},y_{k-2}]$. Then $r(z_k)$  could be the function $r(z_k)=z_k$   that is, an auto-regressive model. There exist also alternative configurations such as a nonlinear  auto-regressive model $r(z_k)=[y_{k}^2,y_{k-1},y_{k-2},y_{k}\cdot y_{k-2} ]$ or any possible combination.}
\end{definition}

\begin{definition}[Linear Prediction]\label{def2} 
For an instant $k$, a forecast of $y_{k+1} \in \mathbb{R}$ can be derived through a linear combination of past data, that is:

\begin{equation}\label{PY}
\begin{array}{cll}
\hat{y}_{k+1}(\Psi) &=& b_Y^T \Psi\\
&=& \displaystyle\sum\limits_{j=1}^{v} \Psi_{j} y_{j}
\end{array}
\end{equation}

where $1\leq v \leq k$ , $\Psi \in \mathbb{R}^{v}$ is a weight vector and $b_Y = [y_{1},\ldots,y_v]^T$.
\end{definition}
When $v=k$, all data is used to forecast $y_{k+1}$.  Then, the forecast error can be explained as the difference between $y_{k+1}$ and the linear prediction $\hat{y}_{k+1}(\Psi)$.

\begin{definition}[Prediction error]\label{def3} It is defined the prediction error $\hat{e}_k(\Psi)$, being $k$ the time instant: \begin{equation} \hat{e}_k(\Psi) = y_{k+1}- \hat{y}_{k+1}(\Psi).
\end{equation}
\end{definition}

Thus, the crux of the matter is how to get not only the weight vector $\Psi$ but also an outer limit of the prediction error. This outer limit is estimated by using the assumed relationship between $z_{j-1}$ and $y_{j}$, with $j= 1,2,\ldots,k$ in expression (\ref{ass1eq1}). Then, a set of past components $z_j$  with $j= 0,1,...,k$ should be available. Section \ref{S3} formulates these key ideas. 

\section{Assumptions}\label{S3}
In this section the assumptions are based on some local affine approximations. In order to construct the proposed predictor, the definition of approximation error is used. This is, the result of using the vectors $r(z_{j-1})$ and $\Phi_k$ to infer $y_j$.

\footnote{The reader should note that the point is to relate the {\it k-th} prediction error $e_k$ and the prediction errors generated by using the {\it k-th} vector of unknown parameters $\Phi_k$ with the {\it i-th} regressors $r(z_i)$, with $i= 0,\ldots,k-1$.}

\begin{definition}[Approximation error] \label{def4} For a vector $\Phi_k$, the approximation error $e_{j-1}$ with the pair $(z_{j-1},y_{j})$ being $j=1,2,...,k$ can be defined as:
\begin{equation}
 e_{j-1} =e_{j-1}(\Phi_k)= y_{j}-r(z_{j-1})^T \Phi_k .
\end{equation}
\end{definition}

From now on the dependency of $e_{j-1}(\Phi_k)$ with $\Phi_k$ is omitted. It should be noted that the value of $\Phi_k$ is unknown. The prediction error $\hat{e}_k(\Psi)$ may be biased by the selected vector $\Psi$.  The theorem \ref{theorem1} suggests an approach to define the prediction error $\hat{e}_k(\Psi)$ as a function of the vector $\Psi$ and the aforementioned approximation errors $e_j$ .

Theorem \ref{theorem1} proposes an expression to characterize the prediction error $\hat{e}_k(\Psi)$ as a function of vector $\Psi$ and approximation errors $e_j$ previously defined.

\begin{theorem}  \label{theorem1} For either vector $\Psi \in \mathbb{R}^{v}$ so that
\begin{equation} \label{exp1}
\displaystyle\sum\limits_{j=1}^{v} \Psi_{j} r(z_{j-1})= r(z_{k}),
\end{equation} 	
\end{theorem}
so the prediction error $\hat{e}_k(\Psi) = y_{k+1}- \hat{y}_{k+1}(\Psi)$ is set as a linear combination of the approximation errors $e_j$, this is
$$\hat{e}_k(\Psi)= -\displaystyle\sum\limits_{j=1}^{v}\Psi_{j} e_{j-1} + e_k.$$
Remark that $\Psi_i$ refers to the j-$th$ item of vector $\Psi$. A proof of the theorem can be found in the Appendix section \ref{Mathderiv}. Matricially, expression (\ref{exp1}) is equivalent to $\Psi \in \{\Psi \;:\; A^T\Psi = r(z_k)\}$ where matrix $A$ is:

\begin{equation}\label{A}A^T = \left[
               \begin{array}{ccccccc}
                r(z_{0}) & r(z_{1}) & ...  & r(z_{v-1}) \\\end{array}
             \right].
\end{equation}

It is necessary to know the vector $\Phi_k$ to get an error value $e_{j-1}$. Alternatively, other properties of $e_{j-1}$ can also be assumed. Both deterministic and stochastic options are available in the literature. In a deterministic view, an upper bound of $|e_{j-1}|$ is considered. This idea is discussed in the section \ref{Det_err}.

\subsection{Deterministic error}\label{Det_err}
In methods with bounded-error \citep{MilaNortPieWal96}, a parametric model and an unknown but bounded-error are regarded. An upper limit of this error is expected to estimate a set of consistent parameters. Similar assumptions are presumed in this work in order to develop a predictor with deterministic assumptions.

\begin{assumption}\label{ASS_3} Constants $\sigma,L \geq 0$ are set such that approximation errors $e_{j-1}$ and $e_k$ are delimited by expressions \begin{equation} |e_{j-1}| \leq\sigma+L||z_{j-1}-z_k||\end{equation} with $j=1,...,k$ and \begin{equation} |e_{k}| \leq\sigma\end{equation} being $||\cdot||$ a norm.
\end{assumption}

The error term is bounded by $|e_k|\leq\sigma$. The assumption \ref{ASS_3} has been broadly used in the  bounded-error system identification's context \citep{MilaNortPieWal96}. Remark that $\sigma$ is the tunning parameter that adds the minimum level of noise considered and $L$ the tunning parameter of uncertainty due to the local affine approximation.

\begin{remark}\label{rmk1}Historical data can be used to estimate an approximate value of $\sigma$ and $L$ when no prior knowledge of these constants is available. In \citep{BraAla2017} a method based on bounded-error and non-counterfeit data is provided.
\end{remark}

\begin{lemma} Considering Assumptions \ref{ASS_1} and \ref{ASS_3}, for any $\Psi$ such that $A^T\Psi=r(z_k)$, prediction error $\hat{e}_k(\Psi) = y_{k+1}- \hat{y}_{k+1}(\Psi)$ is bounded by:
\begin{equation}\label{exp2}|\hat{e}_k(\Psi)|\leq \displaystyle\sum\limits_{j=1}^{v} |\Psi_{j}| (\sigma+L||z_{j-1}-z_k||)+\sigma.
\end{equation}

{\it Proof.}Through a straightforward application of Theorem \ref{theorem1} and bound $|e_i|\leq\sigma+L||z_j-z_k||$ is obtained the expression (\ref{exp2}).
\emph{QED}
\end{lemma}

At this point the possibility of considering how to obtain the vector $\Psi$ is established. A wise option is to use the vector that minimizes an upper bound of  $|\hat{e}_k(\Psi)|$ using the expression (\ref{exp2}).

\begin{definition}[Deterministic predictor] \label{DP}The deterministic prediction $\hat{y}_{k+1}(\Psi^D)$ is defined by $$\hat{y}_{k+1}(\Psi^D) = \displaystyle\sum\limits_{j=1}^{v}\Psi^D_{j}y_{j},$$ where vector $\Psi^D$ adresses the problem of constrained linear optimization as follows
\begin{equation} \label{MIP2}\begin{array}{ccc}
                  \Psi^D =&arg \min\limits_{\Psi} & ||W_k\Psi||_1\\
                 &s.t. & A^T\Psi =  r(z_k)\\
                 \end{array}\end{equation}
where $W_k$ is a diagonal matrix with central items $w^k_{j,j}=\sigma+L||z_{j-1}-z_k||$ with $j=1,...,v$. Then, an upper bound of the absolute value of the prediction error is minimized by the vector $\Psi^D$.
\end{definition}

It is important to note that the notation $\Psi^D$ refers to the deterministic nature of the estimate. Expression (\ref{MIP2}) use $L_1$-norm to obtain the vector solution $\Psi^D$. In this case, $\Psi^D$ is sparse, that is, most of number of components $\Psi^D_i$ of vector $\Psi^D$ are zero. As $\Psi^D$ is a sparse matrix and considering Definition \ref{DP} then it is deduced that $\hat{y}_{k+1}(\Psi^D)$  use a relatively short number of measurements $y_i$. 

\subsection{Stochastic error}
The stochastic view consider the approximation error $e_j$ as a random variable. So there are some assumptions about the mean and the variance of $e_j$. Specifically there are assumptions in the dimension of variance of $e_j$.

\begin{assumption}\label{ASS_4} The independent variables, approximation error $e_{j-1}$ and error term $e_k$ have zero mean and variances bounded by $var(e_{j-1})\leq \sigma + L||z_{j-1}-z_k||$ and $var(e_k) \leq \sigma$ accordingly. Positive values of constants $\sigma$ and $L$ is taken as prior knowledge.
\end{assumption}

As indicated in Remark \ref{rmk1}, if not available previous knowledge of the constants $\sigma$ and $L$, historical data may be used to obtain an estimation. The variance of error $e_{j-1}$ consists of a minimum value defined by $\sigma$ and a term depending of the local approximation, i.e. $||z_{j-1}-z_k||$. it is possible to extend that as $e_{j-1}$ and $e_k$ are random variables then $\hat{e}_k(\Psi)$ is also random and therefore other properties can be derived.

\begin{assumption} Taking into account the previous Assumptions \ref{ASS_1} and \ref{ASS_4}, for any $\Psi$ such that $A^T\Psi=r(z_k)$, prediction error $\hat{e}_k(\Psi) = y_{k+1}- \hat{y}_{k+1}(\Psi)$ is a random variable with zero mean and variance, it is defined by:
\begin{equation}\label{SVAR}\begin{array}{cll}
var(\hat{e}_k(\Psi) )  & = & \displaystyle\sum\limits_{j=\underline{1}}^{v} \Psi_j^2 var(e_{j-1})+ \sigma     \\
&\leq& \displaystyle\sum\limits_{j=1}^{k} \Psi_{j}^2 (\sigma+L||z_{j-1}-z_k||)+\sigma. \\
\end{array}
\end{equation}
\end{assumption}

At this point,it is possible to formulate a predictor that minimize the outer bound of the variance prediction error.

\begin{definition}[Stochastic prediction]
The stochastic prediction $\hat{y}_{k+1}(\Psi^S)$ is defined by: 
$$\hat{y}_{k+1}(\Psi^S) = \displaystyle\sum\limits_{j=1}^{v}\Psi^S_{j}y_{j},$$
being $\Psi^S$ a vector that solves a constrained linear optimization problem as follows:
\begin{equation} \label{MIP}\begin{array}{ccc}
                  \Psi^{S} =&arg \min\limits_{\Psi} & \Psi^TW_k \Psi\\
                 &s.t. & A^T\Psi =  r(z_k).\\
                 \end{array}
\end{equation}
An explicit notation of this optimization problem is:

\begin{equation}\label{LS}\Psi^{S} = W_k^{-1}A(A^TW_k^{-1}A)^{-1}r(z_k).
\end{equation}
\end{definition}

In the same way, $\Psi^{S}$ highlights the stochastic assumptions considered to get the estimate. The following equality is satisfied
$$\hat{y}_{k+1}(\Psi^S) = b_Y^T\Psi^S = r(z_k)^T\Phi^*,$$
where $\Phi^* = (A^TW_k^{-1}A)^{-1}A^TW_k^{-1}b_{Y}$ is the argument of which minimizes the a quadratic prediction-error, with the following  cost function:
\begin{equation}\begin{array}{rl} J(\Phi)= & (b_{Y}-A\Phi)^T W_k^{-1} (b_{Y}-A\Phi)\\
=& \displaystyle\sum\limits_{j=1}^{k}\frac{(y_{j}-r(z_{j-1})^T\Phi)^2}{(\sigma+L||z_{j-1}-z_k||)}.
  \end{array}  \label{Jfun}\end{equation}

In this way, the stochastic prediction is equivalent to solve a weighted least-squares problem where the weights are set by the items of the diagonal of $W_k$ squared. Commonly, $\Psi^{S}$ is not a sparse vector, this is that most items are non zero numbers. So, in order to get the prediction $y_{k+1}$ a great number of $y_j$ would be used.

The goal of this paper is to bring a predictor that combines the two predictions based on the different assumptions obtained from $\Psi^D$ and $\Psi^S$ respectively. Section \ref{secVI} introduces the key points of this paper.

\section{Proposed predictor}\label{secVI}
This work proposes to obtain an estimation of the output $y_{k+1}$ by a linear combination of past data $y_j$, with $j = 1, 2,...,v$ where $v \leq k$ (\cite{RollNazLjung05}). Next, a formal definition of the proposed predictor is provided. This definition use a constant $\gamma \geq 0$ to balance the deterministic or stochastic nature of the prediction.

\begin{definition}\label{definitionCIP}
Given a constant $\gamma \geq0$, the predictor $\hat{y}_{k+1}(\Psi^*)$ is defined by $\hat{y}_{k+1}(\Psi^*) = \displaystyle\sum\limits_{j=1}^{k} \Psi^*_jy_j$ where $\Psi^*$ is the optimal solution of:
\begin{equation} \label{CIP}\begin{array}{ccc}
                  \Psi^*(\gamma) =&arg \min\limits_{\Psi} & ||W_k\Psi||_1\\
                 &s.t. & A^T\Psi = r(z_k)\\
                 && ||\Psi-\Psi^{S}||_1 \leq \gamma
                 \end{array}
\end{equation}
and vector $\Psi^{S}$ is defined in (\ref{LS}).

\end{definition}

Some qualitative properties of the proposed predictor can be clarified. Note that, expression (\ref{CIP}) is a  constrained linear convex optimization problem and can be solved in an efficient way \cite{Boyd04}. Assuming that (\ref{CIP}) has a bounded solution, there is a constant $\bar{\gamma}$ such that if $\gamma \geq \bar{\gamma}$ then equality $\Psi^* = \Psi^D$ is obtained. Term $||\Psi-\Psi^S||_1$ of expression (\ref{CIP}) takes into account the stochastic Assumption explained in Section \ref{ASS_4} to obtain the optimal solutions $\Psi^*$. If $\gamma=0$ then $\Psi^* = \Psi^S$. So, constant $\gamma$ can be seen as a tuning parameter to balance the deterministic or stochastic nature of the considered approximation error.

\begin{remark}It is important to remark that the proposed predictor encompasses some relevant nonparametrics predictors. If $\gamma=0$ and $r(z_k)=1$ the proposed predictor is equivalent to the Nadaraya-Watson predictor \cite{Narayada64,Wats:1964}. On the other hand if $\gamma = 0$ and $r(z_k)=[z_k^T \; 1]$ a predictor based on Local Linear Regression is obtained. Besides, if $\gamma = 0$ and if $L = 0$ a parametric auto-regressive linear regression is performed.
\end{remark}

\begin{remark} It is important to remark that following similar reasoning it is possible to obtain different forecasting horizons. This is represented by the expression $y_{k+h}(\Psi^{*})$ being $h \geq 1$ the number of steps ahead.
\end{remark}

\section{Results}\label{ejemplo}
In this section results are shown, the predictor was performed in four time series: the Monthly airline passenger numbers, the Canadian lynx data, the Monthly critical radio frequencies in Washington, D.C. and the Monthly pneumonia and influenza deaths time series are used in order to demonstrate the appropriateness and effectiveness of the proposed predictor. These time series come from different areas and have different statistical properties, so is a suitable benchmark to test time series predictors.

Subsection \ref{sub_considerations} explain the characteristics of the study performed, hyper-parameterization, kernels, error measures and more details are exposed below. 

\subsection{Considerations}\label{sub_considerations}
\begin{itemize}
\item To simplify the study, the proposed predictor (denoted $CP$) is considered with values $\sigma = 0$ and $L = 1$ in all cases. Note that $\sigma$ and $L$ could be considered hyper-parameters in order to improve the results obtained by the proposed predictor in this study.

\item The proposed predictor ($CP$) is compared to three Nadaraya-Watson predictors (denoted as $NW1$, $NW2$ and $NW3$) using Epanechnikov, Gaussian and Tricube kernel functions respectively and three local linear regression models (denoted as $LL_1$, $LL_2$ and $LL_3$) using Epanechnikov, Gaussian and Tricube kernel functions respectively to define the local weights. Table \ref{KF} shows the expression of weights $w_{i,i}$ with $i=1,...,N$ for the aforementioned kernel functions. A bandwidth $\gamma$ is considered in the non-parametric predictors. Also, an auto-regressive linear regression $(AR)$ is indirectly included in the benchmark, as the proposed predictor also includes this model according to the hyperparameter combinations as explained in remark 2 of section  \ref{definitionCIP}.

\begin{table}[ht]
\caption{Kernel functions}
\label{KF}
\begin{center}
\begin{tabular}[c|p{0.1cm}]{|c|c|c|c|c|c|}
\hline   Epanechnikov& Gaussian  & Tricube   \\
\hline
  $w_{i,i} = \left \{ \begin{array}{cc} 1-v_i^2 &if \; |v_i|\leq 1 \\   0  &if \; |v_i|> 1 \end{array} \right.$ &
  $w_{i,i}=e^{-\frac{1}{2}v_i^2}$   &
  $w_{i,i} = \left \{ \begin{array}{cc} (1-|v_i|^3)^3  &if \; |v_i|\leq 1 \\   0  &if \; |v_i|> 1 \end{array} \right.$ \\
\hline
\multicolumn{3}{|c|}{ $v_i = \frac{||z_i-z_k||}{\gamma}$} \\
\hline
\end{tabular}
\end{center}
\end{table}

\item Two different forecast consistency measures are used in order to compare the predictor performances with the aforementioned  models: Mean Absolute Error (MAPE) and Symmetric Mean Absolute Error (SMAPE) that have been studied by several authors \citep{Armstrong85}. Mean Absolute Error is defined by

\begin{equation} \label{MAPE}\begin{array}{ccc}
                  MAPE = \frac{100}{n}\displaystyle\sum_{t=1}^n  \left|\frac{y_t-\hat{y}_{t}}{y_t}\right|,
                 \end{array}
\end{equation}

where $\hat{y}_{t}$ and ${y_t}$ are the predicted and observed data, respectively, and $n$ is the number of data. The second criterion is the Symmetric Mean Absolute Percentage error (SMAPE), which is

\begin{equation} \label{SMAPE}\begin{array}{ccc}
                  SMAPE = \frac{100}{n} \displaystyle\sum_{t=1}^n \frac{\left|\hat{y}_{t}-y_t\right|}{(|y_t|+|\hat{y}_{t}|)/2},
                 \end{array}
\end{equation}

where $\hat{y}_{t}$ and ${y_t}$ are the predicted and observed data, respectively, and $n$ is the number of data.

\item In order to train the predictors, the time series has been splitted into a training and test set with a percentage between 70-90 for a training set and a percentage between 10-30 of data for a test set. The aforementioned error measures described in formulas (\ref{MAPE}) and (\ref{SMAPE}) are selected to benchmark the predictors, not only because of its interpretability but also because are scale-independent \citep{Robscaleindependent}. The training set, a leave-one-out cross validation approach, and a grid-search in the hyper-parameter-$\gamma$ space are used to find a suitable value of hyper-parameter $\gamma$. This value is used in the test set to evaluate the prediction methods. 

\item As explained in Section \ref{secVI}, the estimation can be used by a linear combination of the past data, in this way the different models are evaluated by using only the training set to infer the prediction this is when past data $y_j$ is used with $v < k$. Besides, three different forecast horizons are computed by predictor (1 step-ahead, 2 step-ahead and 3 step-ahead).
\end{itemize}

In order to test the proposed predictor, two different type of data are used to perform the benchmark, the first in subsection \ref{sub_acad} where some famous and more academic time series are used to compare the models, the second in subsection \ref{electricity} a real life time series is used to test the predictor.

\subsection{Academic Time series}\label{sub_acad}
This subsection compare 6 non-parametric models against the proposed model in four different time series, performing forecasts in three different predicting horizon lengths. Each time series provides the results in bar-plots by $MAPE$ and $SMAPE$ errors in the test set for the proposed forecasting horizons. A final sub-subsection averages all time series results in order to extract general conclusions for the different benchmarked models.

\subsubsection{Airline passengers dataset}\label{sub_air}
The classic Box and Jenkins airline data contains monthly totals of international airline passengers from $1949$ to $1960$ \citep{BoxJen76}.

\begin{figure}[!ht]
\begin{subfigure}{0.5\textwidth}
\includegraphics[width=1\linewidth, height=60mm]{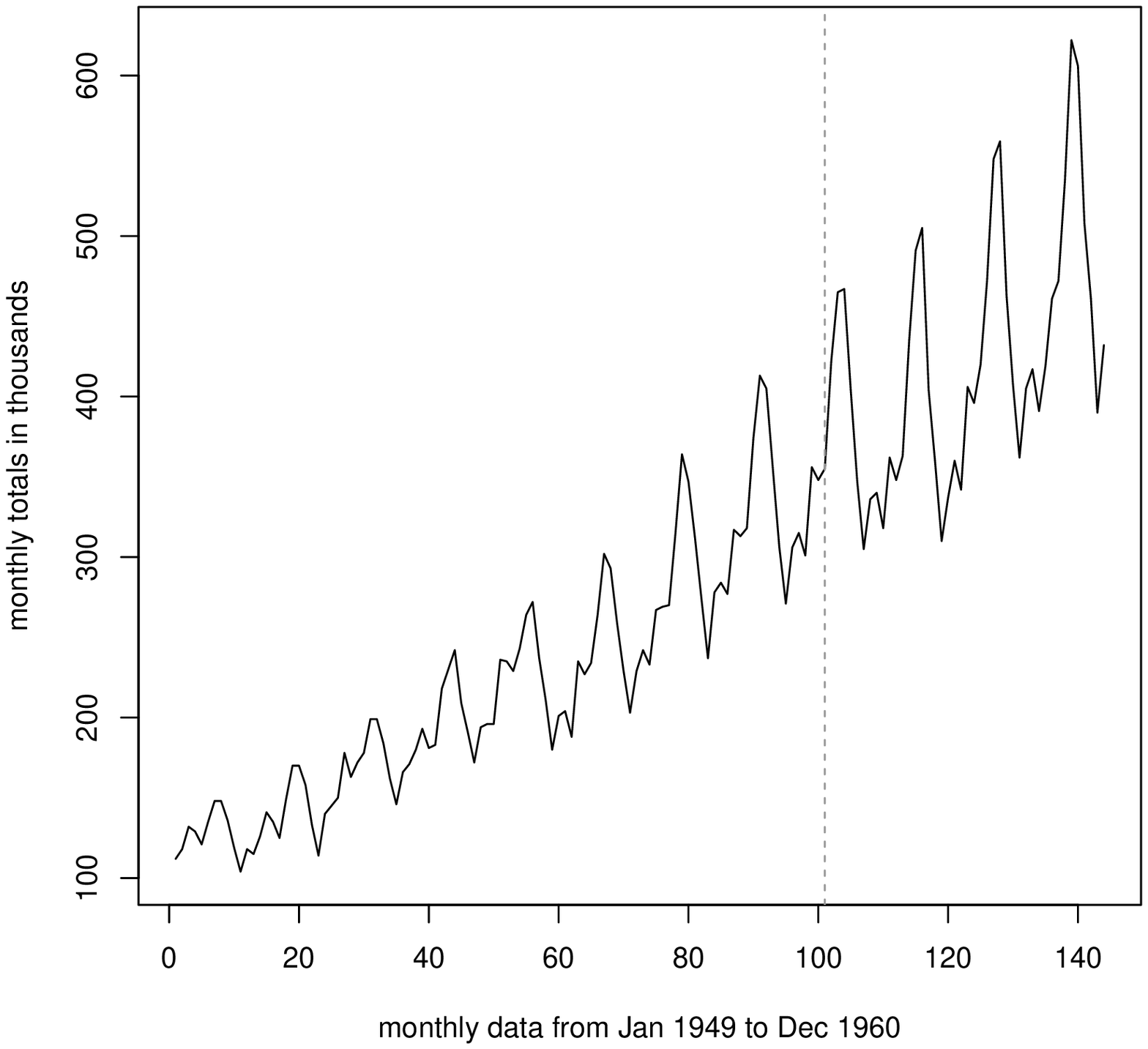}
\vspace*{-4mm}
\caption{Original Time series}
\label{Airpassengers.eps}
\end{subfigure}
\begin{subfigure}{0.5\textwidth}
\includegraphics[width=1\linewidth, height=60mm]{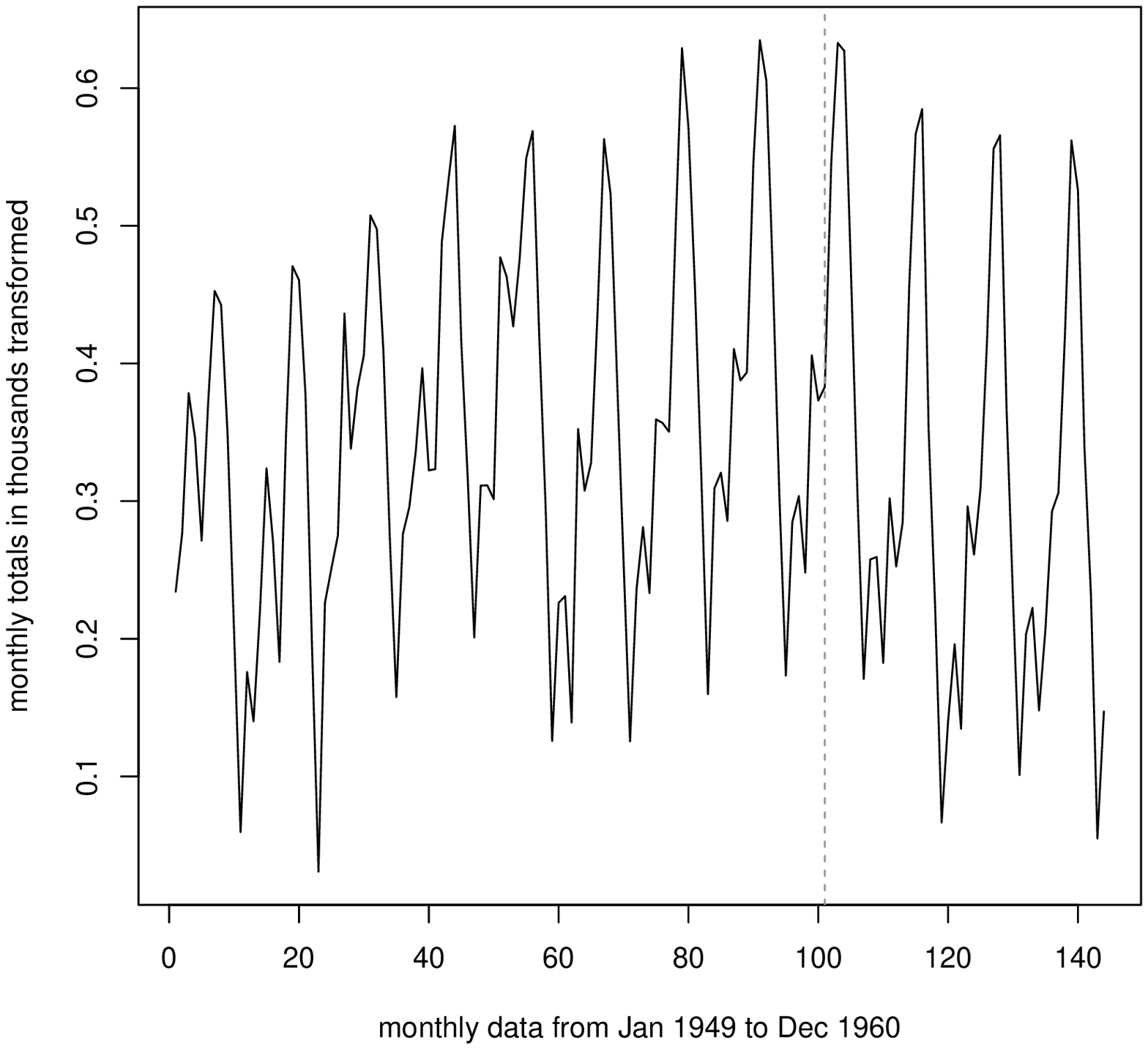}
\vspace*{-4mm}
\caption{Transformed Time series}
\label{Airpassengers_transformed.eps}
\end{subfigure}
\caption{Monthly totals of international airline passengers (1949 $-$ 1960).}
\label{fig:air_pred_NR}
\end{figure}

This time series plotted in Figure \ref{Airpassengers.eps} has $144$ observations, the first $101$ observations were used as training set and the last $43$ as test set. It has also been very analyzed in the time series literature.

As explained in Section \ref{SecForm}, it is considered a detrended time series. In this sense a log with base 10 and a linear detrend function are applied to transform the data, it is shown the data set transformed in Figure \ref{Airpassengers_transformed.eps}.

\begin{figure}[!ht]
\centering
\includegraphics [width=70mm,height=60mm]{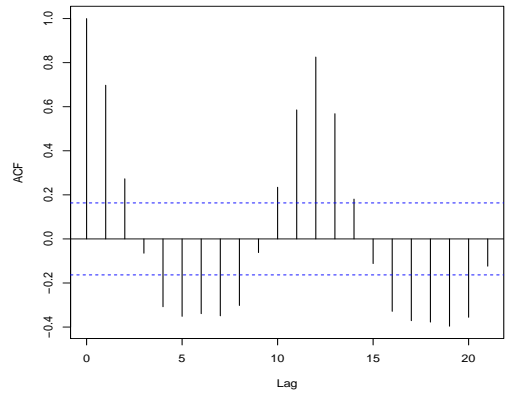}
\vspace*{-4mm}
\caption{Auto-correlation function of Airline passengers transformed time series.}
\label{acf_air.eps}
\end{figure}

The auto-correlation plot of the transformed data set shown in figure \ref{acf_air.eps} at the appendix represents shows a high correlation between observations of this time series that are separated by $k=12$ time units, in this sense, the predictor can be represented as $r(z_k)= z_k = [y_{k-1} \;y_{k-2}\;...\;y_{k-12}]^T$.

\begin{figure}[!ht]
\centering
\includegraphics [width=90mm,height=70mm]{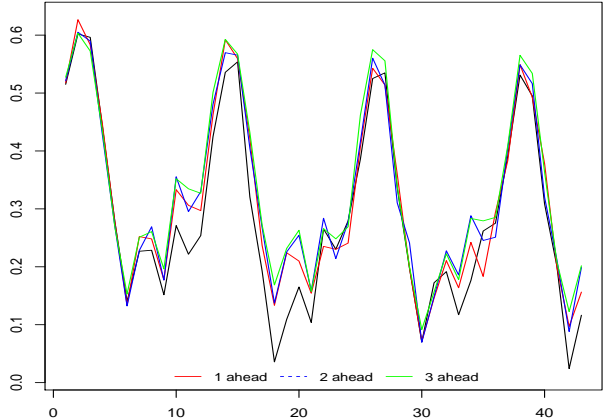}
\vspace*{-4mm}
\caption{International airline passengers predictions by forecasting horizon in the test set.}
\label{air_pred_MAPE_R.eps}
\end{figure}

Figure \ref{air_pred_MAPE_R.eps} shows the forecasts of the proposed predictor by forecasting horizons with the hyper-parameters selected in both error measures.

The hyper-parameter $\gamma$ is selected in the training set where the error is minimum, the value of $\gamma$ is inferred to perform forecasts in the test set. Besides, depending on the error measure selected in the training set the results may vary, in this case, same optimal hyper-parameters are found in both error measures, these results are in table \ref{table_optim_air}.

\begin{table}[!ht]
\caption{Airline passengers time series optimal gamma.}
\label{table_optim_air}
\centering
\begingroup\fontsize{8pt}{8pt}\selectfont
\begin{tabular}{lrrr}
  \hline
Ahead & $\gamma$\_mape & $\gamma$\_smape\\ 
  \hline
1.00 & 0.12 & 0.12 \\ 
2.00 & 0.14 & 0.14 \\ 
3.00 & 0.00 & 0.00 \\ 
   \hline
\end{tabular}
\endgroup
\end{table}

\begin{figure}[!ht]
\begin{subfigure}{0.5\textwidth}
\includegraphics[width=1.01\linewidth, height=50mm]{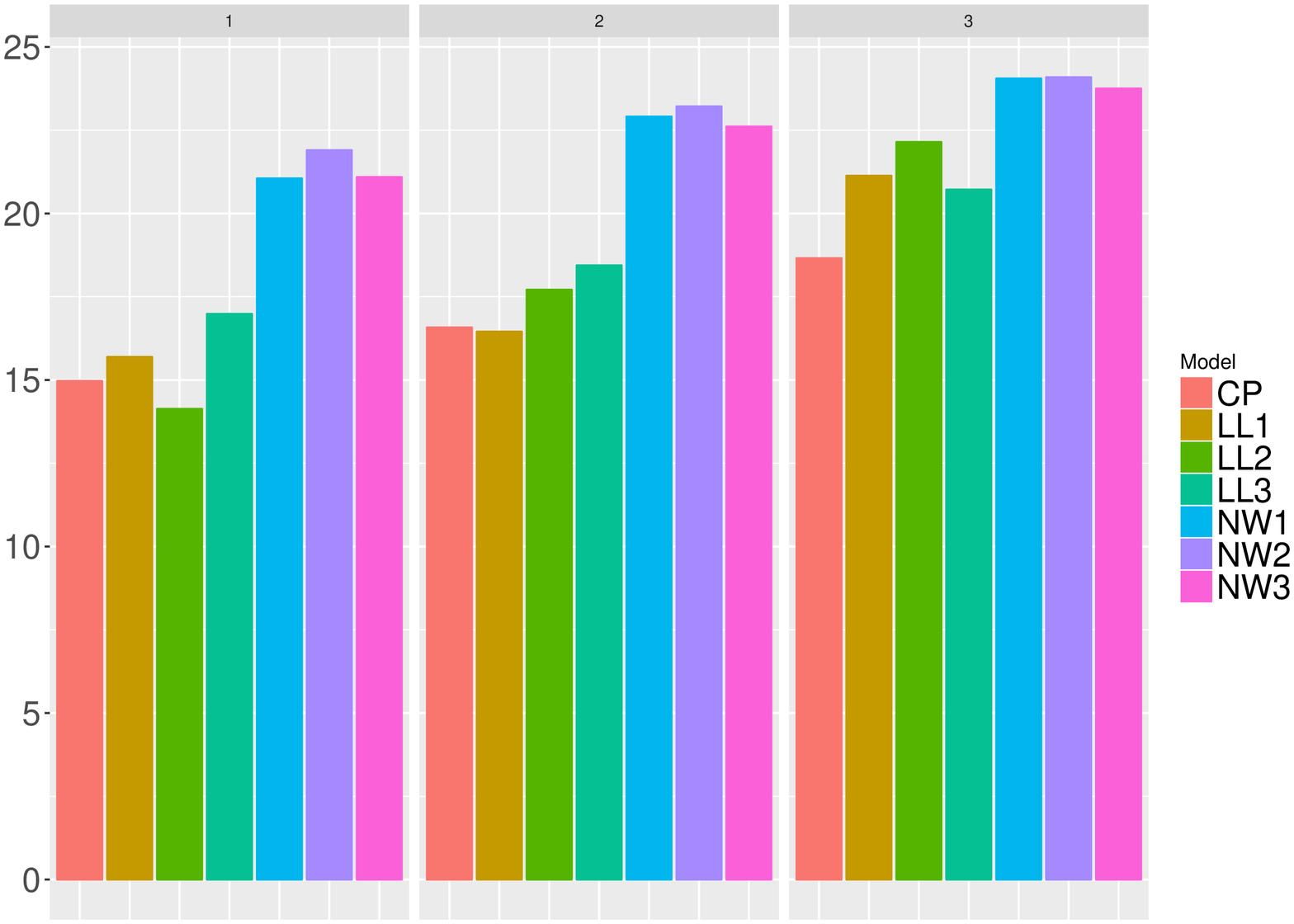}
\vspace*{-4mm}
\caption{SMAPE}
\label{Air_SMAPE.eps}
\end{subfigure}
\begin{subfigure}{0.5\textwidth}
\includegraphics[width=1.01\linewidth, height=50mm]{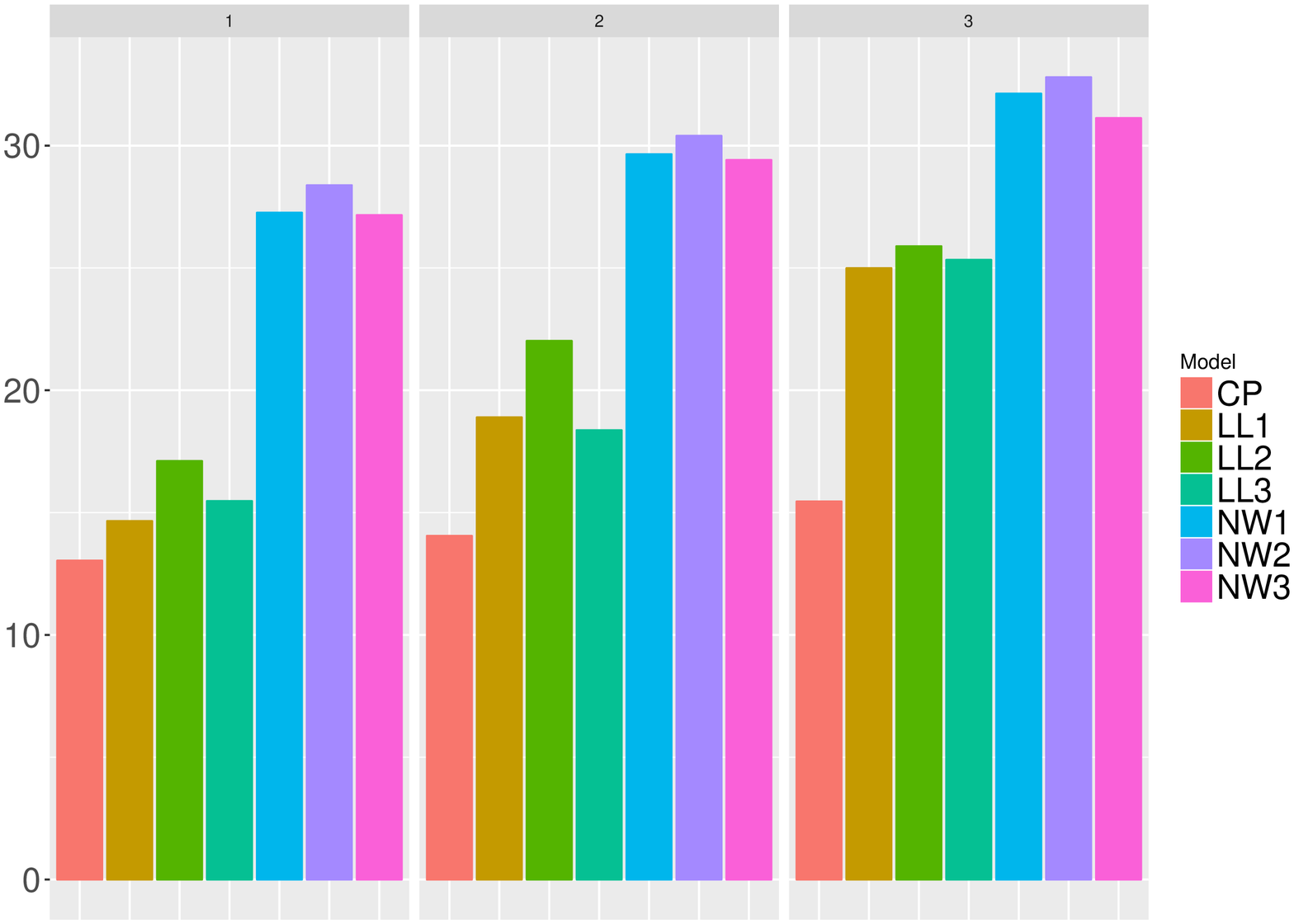}
\vspace*{-4mm}
\caption{MAPE}
\label{Air_MAPE.eps}
\end{subfigure}
\caption{Mean of errors by forecasting horizon in airline passengers time series in test set.}
\label{fig:barplot_air}
\end{figure}

Results of this time series are shown on Table \ref{table:Air_table} in the appendix. To sum up the aforementioned table in a graphical way, figure \ref{fig:barplot_air} show the error measures by predictor and prediction horizon.

Results show by forecasting horizon that the proposed predictor outperforms in both error measures in the proposed forecasting horizons, with the exception of $SMAPE$ error criteria in the one and two step-ahead prediction, where the prediction is close to $LL$ results. Besides, results show that there is not a significant variation of results with the selection of the different kernels.

\subsubsection{Canadian Lynx}\label{sub_canadian}
The Annual numbers of lynx trappings in Canada, contains the number of lynx trapped per year in the Mackenzie River district of Northern Canada from $1821$ to $1934$ \citep{Campbel77}.

It has been extensively analyzed in the time series literature with a focus on the nonlinear modeling. The lynx series plotted in Figure \ref{Lynx.eps} shows a periodicity of approximately 10 years. The lynx series was studied by many researchers found the best-fitted model is AR(12) model \citep{Zhang2003}. In this way, the predictor is based on a auto-regressive model of order $p = 12$, this is $r(z_k)= z_k = [y_{k-1} \;y_{k-2}\;...\;y_{k-12}]^T$.

The lynx series plotted in Figure \ref{Lynx.eps} has $114$ observations, the first 80 observations of this data set were used as training set and the last 34 as test set.

A log with base 10 was applied to the series in order to make a symmetrical data set, the plot of the data set is at Figure \ref{lynx_transformed.eps}:

\begin{figure}[!ht]
\begin{subfigure}{0.5\textwidth}
\includegraphics[width=1\linewidth, height=60mm]{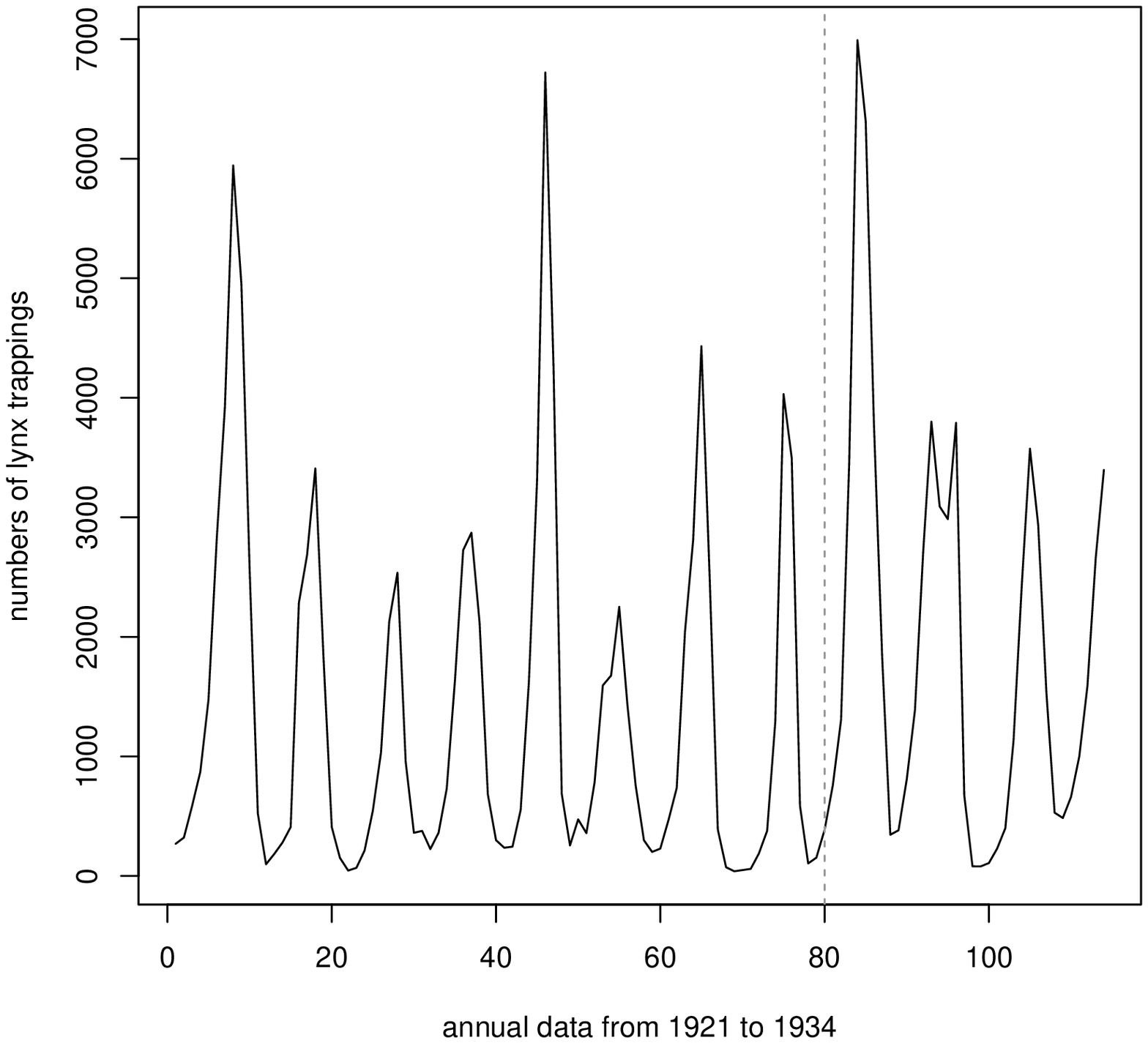}
\vspace*{-4mm}
\caption{Original Time series}
\label{Lynx.eps}
\end{subfigure}
\begin{subfigure}{0.5\textwidth}
\includegraphics[width=1\linewidth, height=60mm]{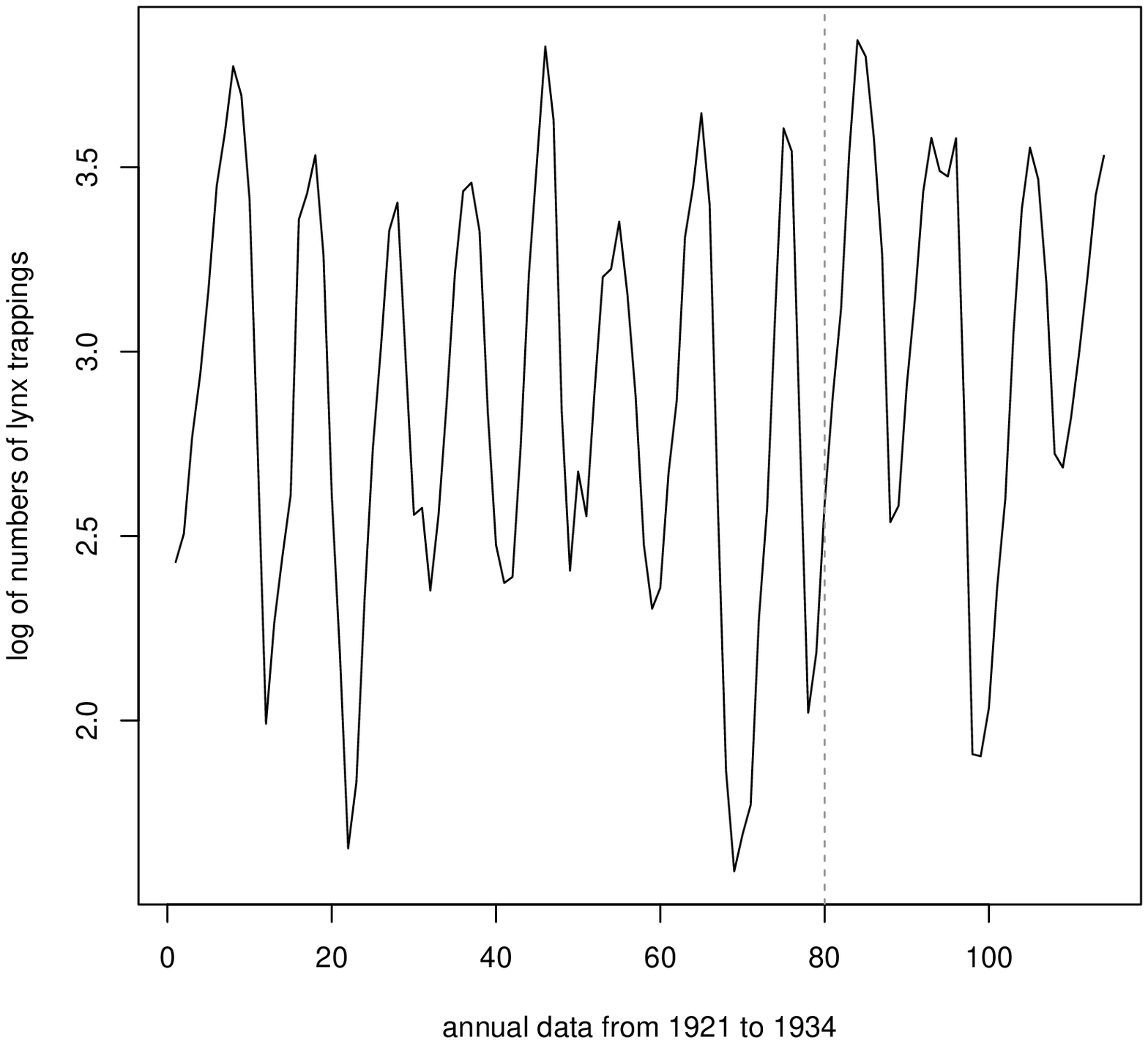}
\vspace*{-4mm}
\caption{Transformed Time series}
\label{lynx_transformed.eps}
\end{subfigure}
\caption{Annual number of lynx trappings in Canada from 1821 to 1934.}
\label{fig:lynx_pred_NR1}
\end{figure}

Figure \ref{fig:lynx_pred_NR1} shows the forecasts of the proposed predictor by forecasting horizons with the hyper-parameters selected in both error measures, in this case, different optimal hyper-parameters are found in the error measure selection on the training set.

\begin{figure}[!ht]
\begin{subfigure}{0.5\textwidth}
\includegraphics[width=1.01\linewidth, height=60mm]{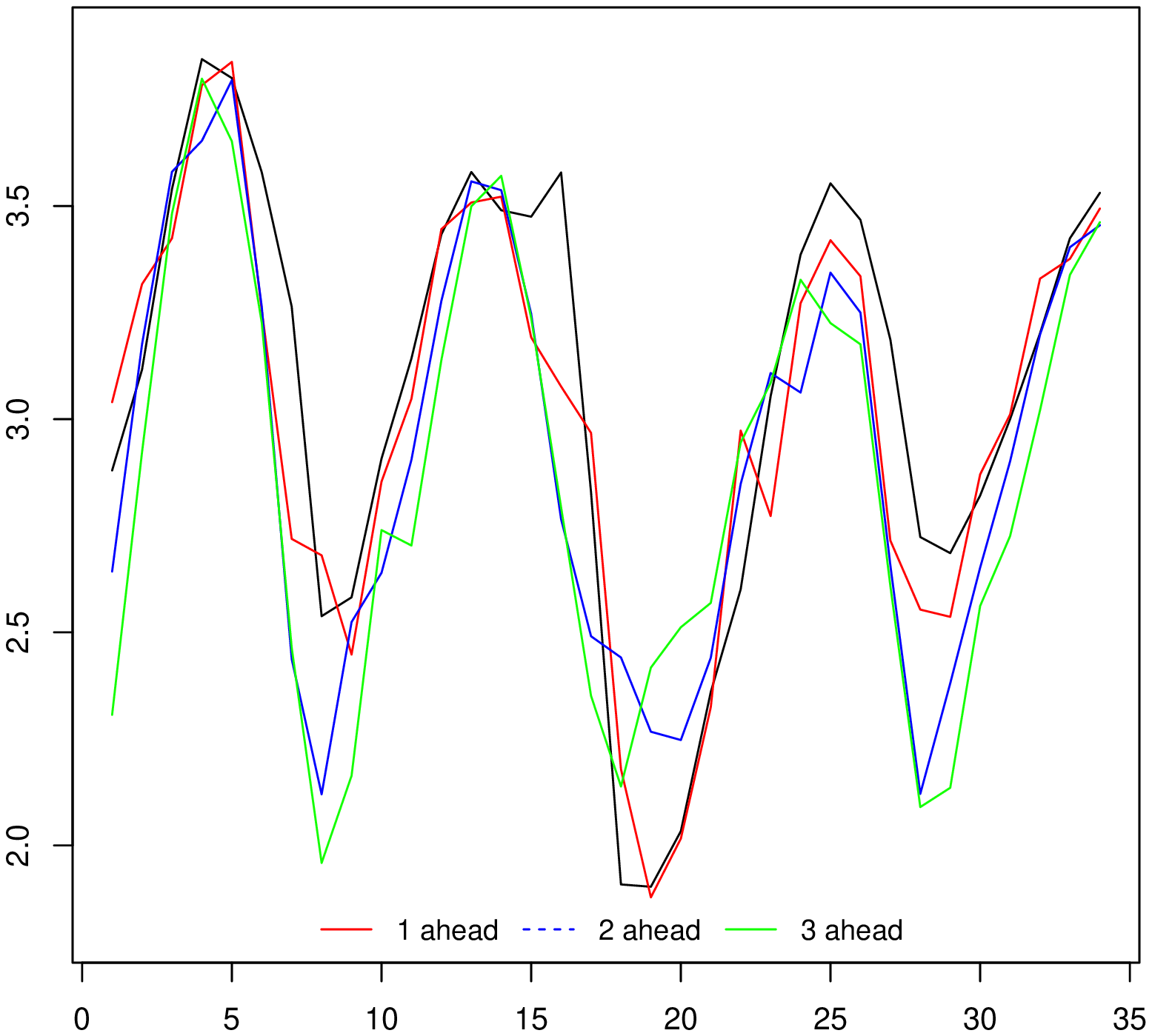}
\vspace*{-4mm}
\caption{MAPE selection criterion}
\label{lynx_pred_MAPE_NR.eps}
\end{subfigure}
\begin{subfigure}{0.5\textwidth}
\includegraphics[width=1.01\linewidth, height=60mm]{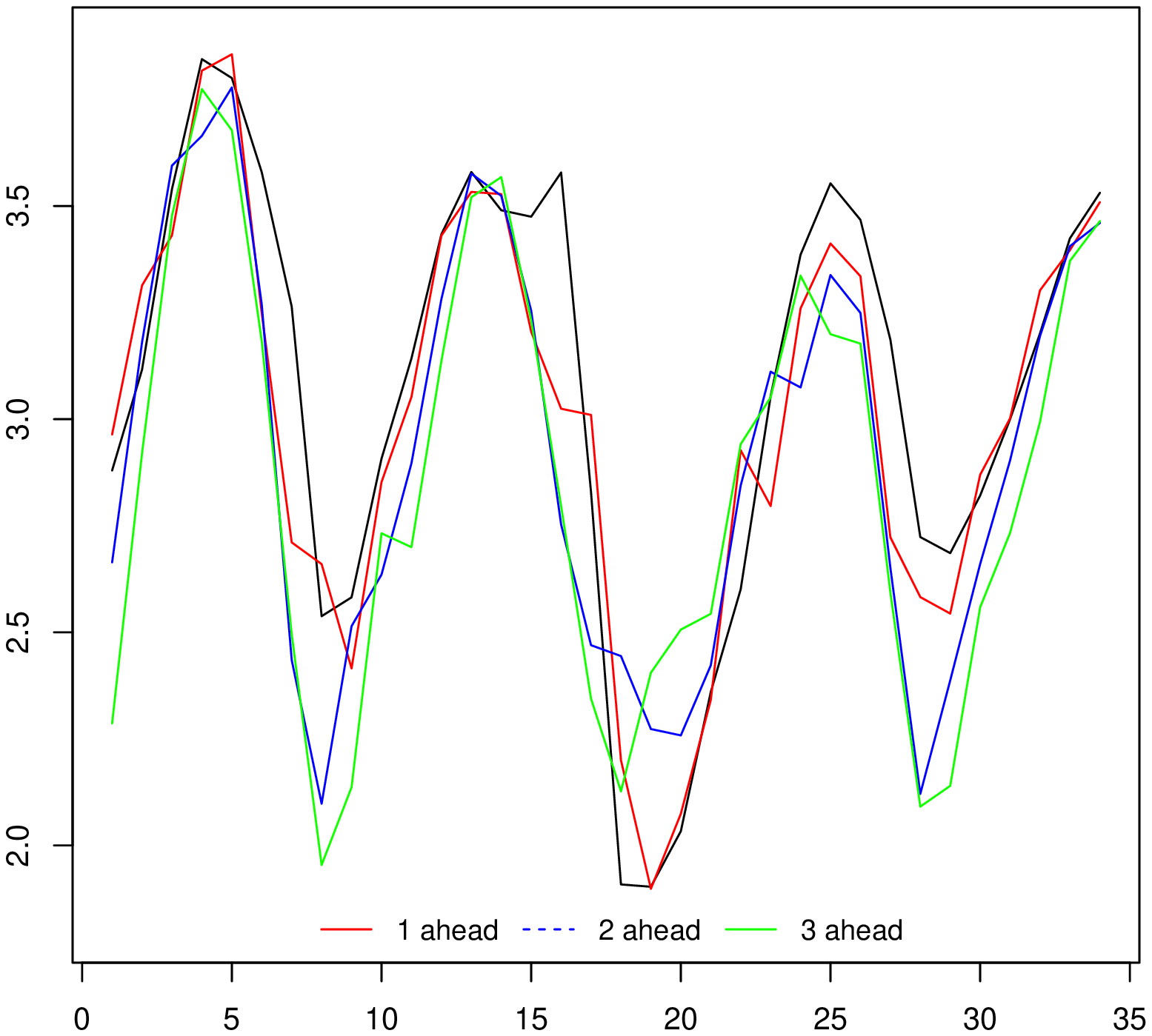}
\vspace*{-4mm}
\caption{SMAPE selection criterion}
\label{lynx_pred_SMAPE_NR.eps}
\end{subfigure}
\caption{Canadyan lynx time series predictions by forecasting horizon in the test set.}
\label{fig:lynx_pred_NR2}
\end{figure}

Table \ref{table_gamma_lynx} shows results that comes from different optimal gamma selections by different error criteria and forecasting horizons.

\begin{table}[!ht]
\caption{Canadian Lynx time series optimal gamma.}
\label{table_gamma_lynx}
\centering
\begingroup\fontsize{8pt}{8pt}\selectfont
\begin{tabular}{lrrr}
  \hline
Ahead & $\gamma$\_mape & $\gamma$\_smape\\  
 \hline
1.00 & 0.01 & 0.06 \\ 
2.00 & 0.00 & 0.02 \\ 
3.00 & 0.02 & 0.04 \\ 
   \hline
\end{tabular}
\endgroup
\end{table}

\begin{figure}[!ht]
\begin{subfigure}{0.5\textwidth}
\includegraphics[width=1.01\linewidth, height=60mm]{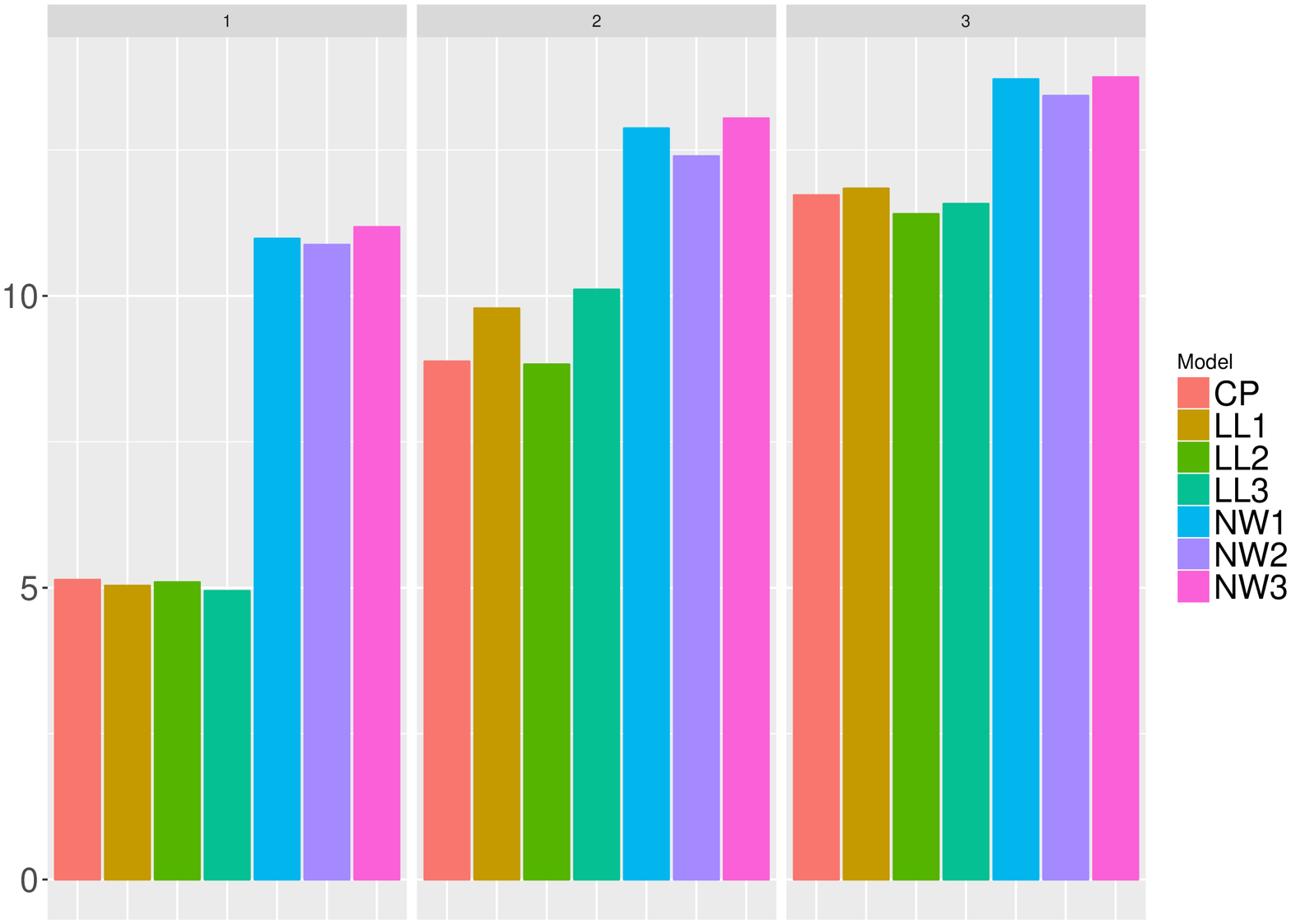} 
\vspace*{-4mm}
\caption{SMAPE}
\label{lynx_SMAPE.eps}
\end{subfigure}
\begin{subfigure}{0.5\textwidth}
\includegraphics[width=1.01\linewidth, height=60mm]{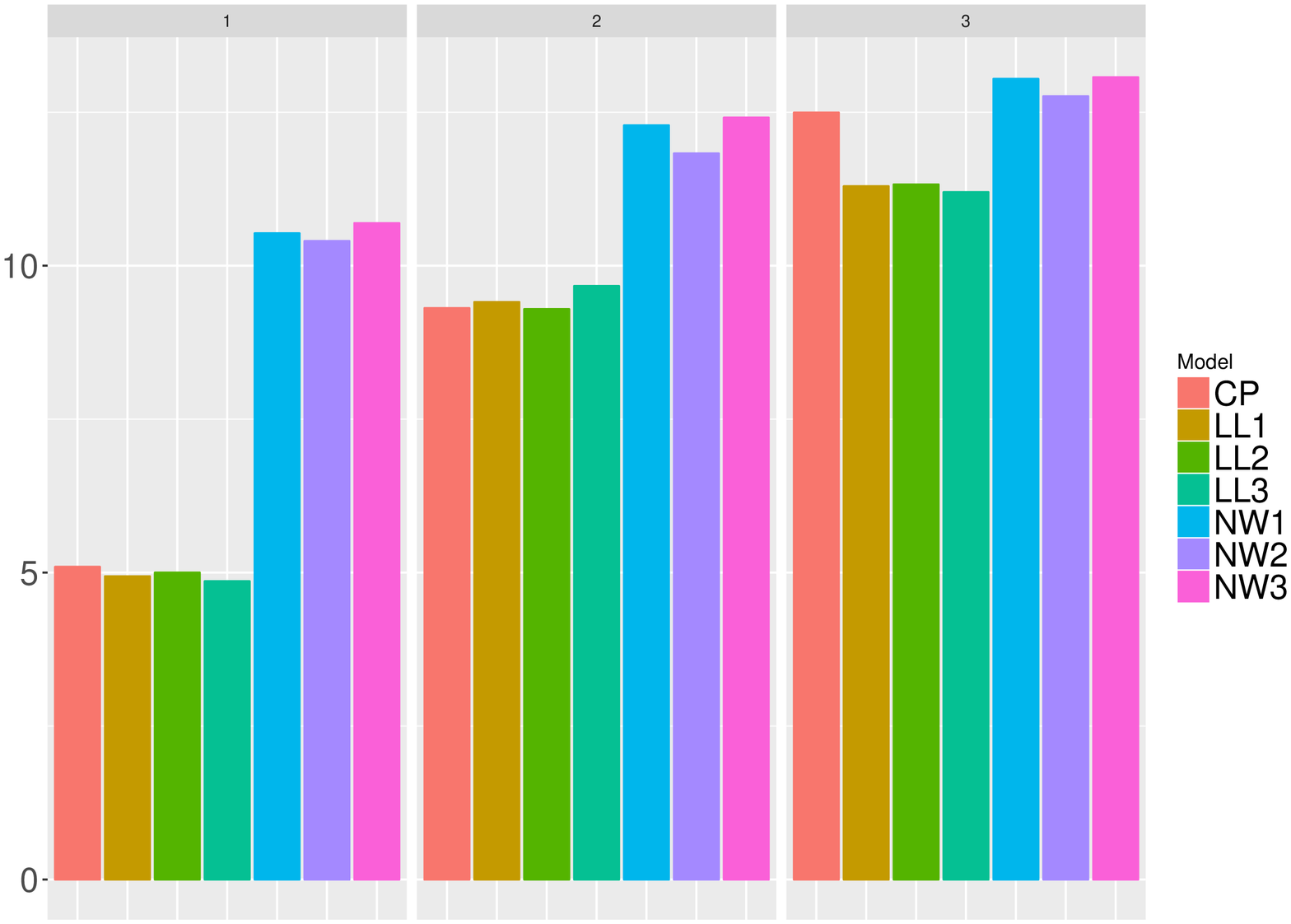}
\vspace*{-4mm}
\caption{MAPE}
\label{lynx_MAPE.eps}
\end{subfigure}
\caption{Mean of test set errors by forecasting horizon in Canadian lynx time series.}
\label{fig:barplot_lynx}
\end{figure}

Corresponding to this predictions, the error measures are shown on Table \ref{table:lynx_table} in the appendix. To sum up this table in a graphical way, the figure \ref{fig:barplot_lynx} plot the error measures by predictor and prediction horizon.

Results in figure \ref{fig:barplot_lynx} show that the proposed predictor get similar results to Local Linear regression in both error measures in the two first proposed forecast horizons, in this case a well-selected kernel on Local Linear regression could be a good option to take into account in order to get closer results in a short prediction term to the proposed predictor $CP$, in a three step-ahead prediction horizon the results mark a tie between $LL$ and $CP$ in the $SMAPE$, on the contrary by selecting $MAPE$ criterion the $LL$ outperforms.

\subsubsection{Monthly critical radio frequencies}\label{sub_radio}
Monthly critical radio frequencies in Washington, D.C., contains the highest radio frequency that can be used for broadcasting from May $1934$ to April $1954$ \citep{DATAMARKET:2014}.

\begin{figure}[!ht]
\centering
\includegraphics [width=90mm,height=60mm]{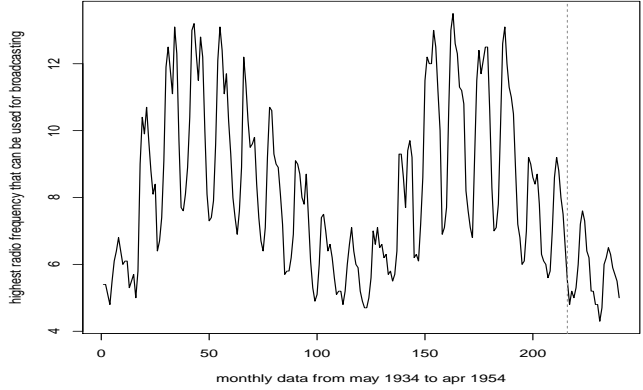}
\caption{Monthly critical radio frequencies (1934$-$1954).}
\label{Radiofrequencies.eps}
\end{figure}

This time series plotted in Figure \ref{Radiofrequencies.eps} has $240$ observations, the first $216$ observations were used as training set and the last $24$ as a test set.
 
\begin{figure}[ht]
\centering
\includegraphics [width=70mm,height=60mm]{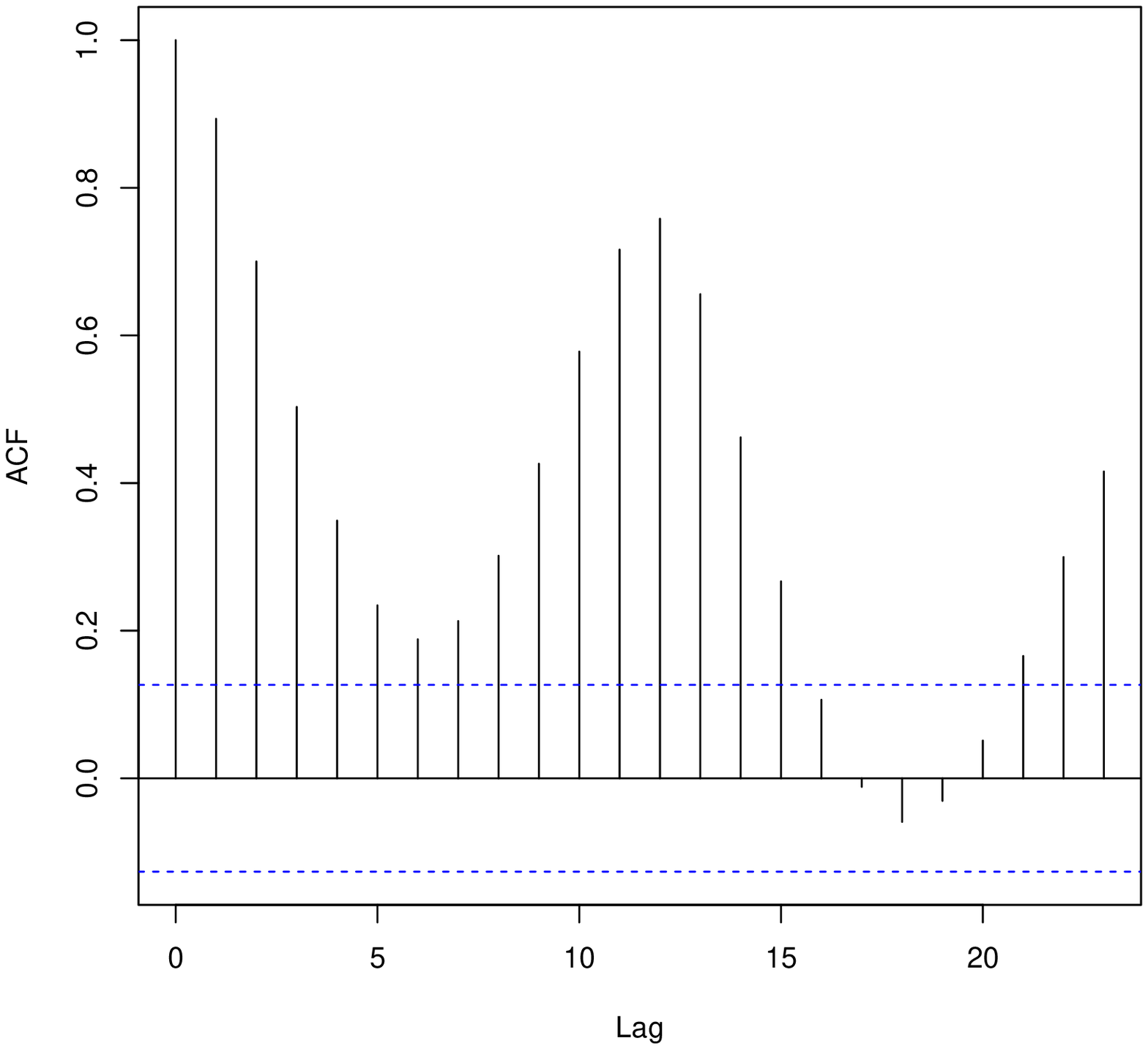}
\caption{Auto-correlation function of Monthly critical radio frequencies time series.}
\label{acf_rad.eps}
\end{figure}

According to auto-correlation plot attached in the Figure \ref{acf_rad.eps} at the appendix, the established model is based on an auto-regressive model of order twelve, which has also been used by many researchers  \citep{RAOANDGABR, Zhang2003}. The auto-regressive model has the shape like $r(z_k)= z_k = [y_{k-1} \;y_{k-2}\;...\;y_{k-12}]^T$.

\begin{figure}[!ht]
\begin{subfigure}{0.5\textwidth}
\includegraphics[width=1.01\linewidth, height=60mm]{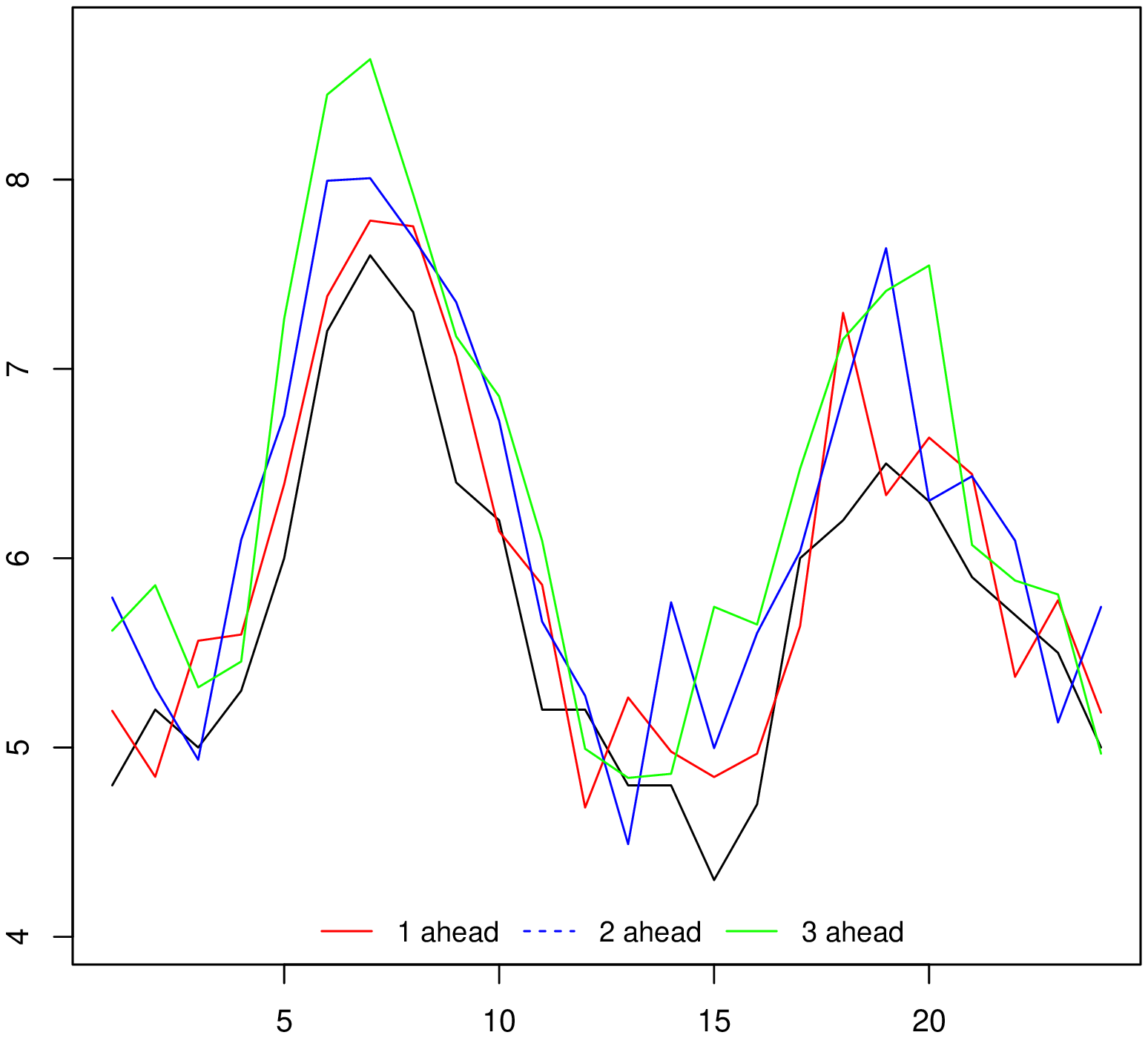}
\vspace*{-4mm}
\caption{MAPE selection criterion}
\label{rad_pred_MAPE_NR.eps}
\end{subfigure}
\begin{subfigure}{0.5\textwidth}
\includegraphics[width=1.01\linewidth, height=60mm]{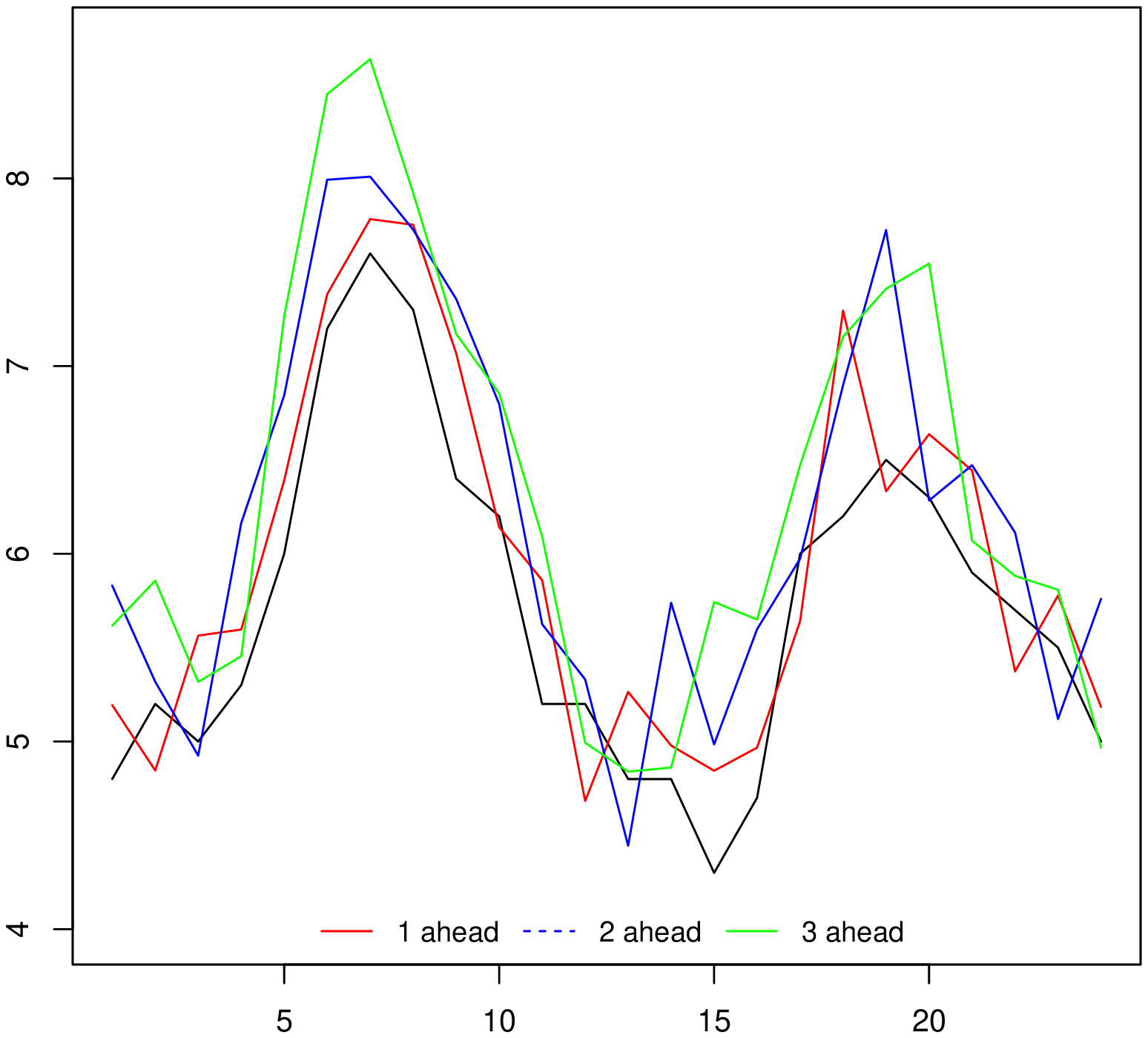}
\vspace*{-4mm}
\caption{SMAPE selection criterion}
\label{rad_pred_SMAPE_NR.eps}
\end{subfigure}
\caption{Monthly critical radio frequencies time series prediction by forecasting horizon in the test set.}
\label{fig:rad_pred_NR}
\end{figure}

Prediction are shown in figures \ref{fig:rad_pred_NR} for the proposed predictor by forecasting horizons with the hyper-parameters selected in both error measures, in this case, different optimal hyper-parameter are found in the different error measure criteria selected on the training set.

Prediction error results of this time series are recorded on Table \ref{table:Rad_table} in the appendix, table \ref{table_gamma_radio} shows the optimal selected gamma. 

\begin{table}[!ht]
\caption{Monthly critical radio frequencies time series optimal gamma.}
\label{table_gamma_radio}
\centering
\begingroup\fontsize{8pt}{8pt}\selectfont
\begin{tabular}{lrrr}
  \hline
Ahead & $\gamma$\_mape & $\gamma$\_smape\\ 
  \hline
1.00 & 0.00 & 0.00 \\ 
2.00 & 0.03 & 0.00 \\ 
3.00 & 0.00 & 0.00 \\ 
   \hline
\end{tabular}
\endgroup
\end{table}

\begin{figure}[!ht]
\begin{subfigure}{0.5\textwidth}
\includegraphics[width=1.01\linewidth, height=60mm]{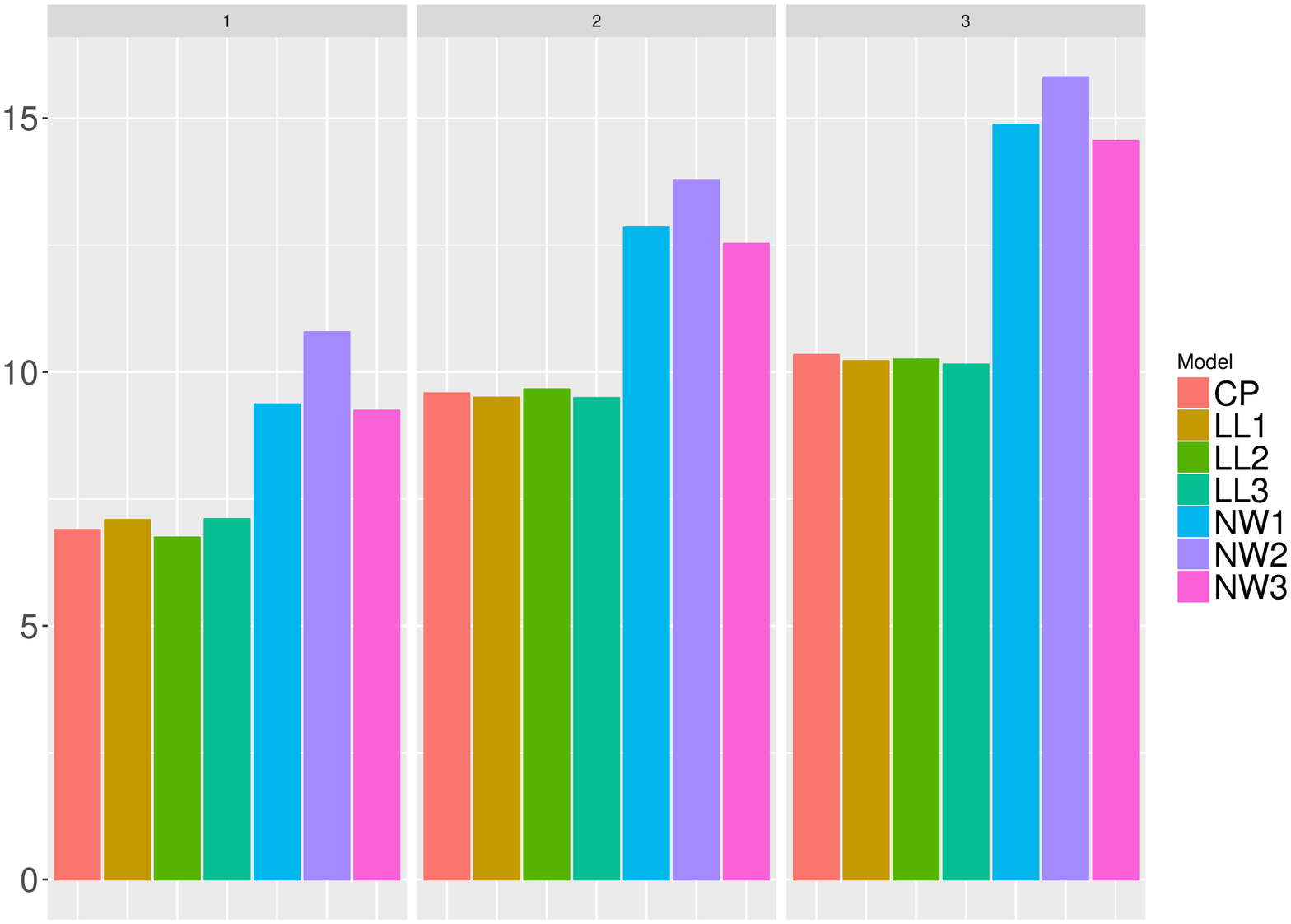}
\vspace*{-4mm}
\caption{SMAPE}
\label{Rad_SMAPE.eps}
\end{subfigure}
\begin{subfigure}{0.5\textwidth}
\includegraphics[width=1.01\linewidth, height=60mm]{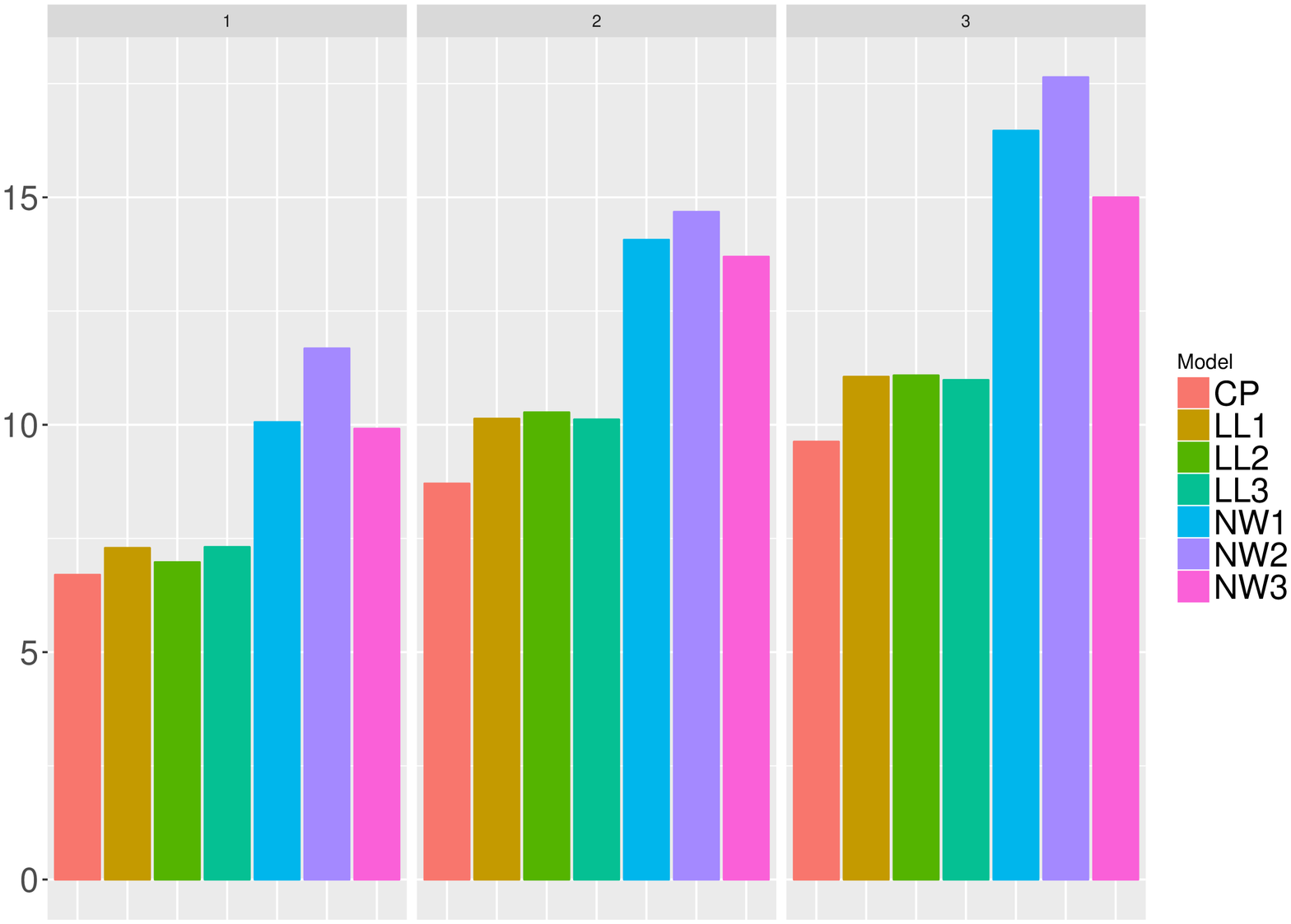}
\vspace*{-4mm}
\caption{MAPE}
\label{Rad_MAPE.eps}
\end{subfigure}
\caption{Mean of test set errors by forecasting horizon in monthly critical radio frequencies time series.}
\label{fig:barplot_rad}
\end{figure}

To sum up the table \ref{table:Rad_table} in a graphical way, figure \ref{fig:barplot_rad} show the error measures by predictor and forecasting horizon.

The proposed predictor get similar results to Local Linear regression in SMAPE measure in the proposed forecast horizons, on the contrary by selecting $MAPE$ measure the Proposed predictor $CP$ outperforms. Besides, results shows that there is not a high significant variation of results with the selection of the different kernels in Local Linear Regression.

\subsubsection{Monthly pneumonia and influenza deaths}\label{sub_Influenza}
Monthly pneumonia and influenza deaths per $10.000$ people in the United States for 11 years, $1968$ to $1978$. 

\begin{figure}[!ht]
\centering
\includegraphics [width=90mm,height=60mm]{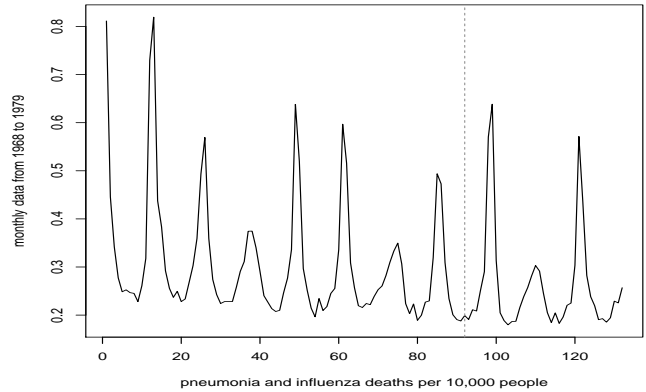}
\caption{Monthly pneumonia and influenza deaths (1968$-$1978).}
\label{flu.eps}
\end{figure}

This time series plotted in Figure \ref{flu.eps} has $132$ observations, the first $84$ observations were used as training set and the last $24$ as a test set.

\begin{figure}[!ht]
\centering
\includegraphics [width=70mm, height=60mm]{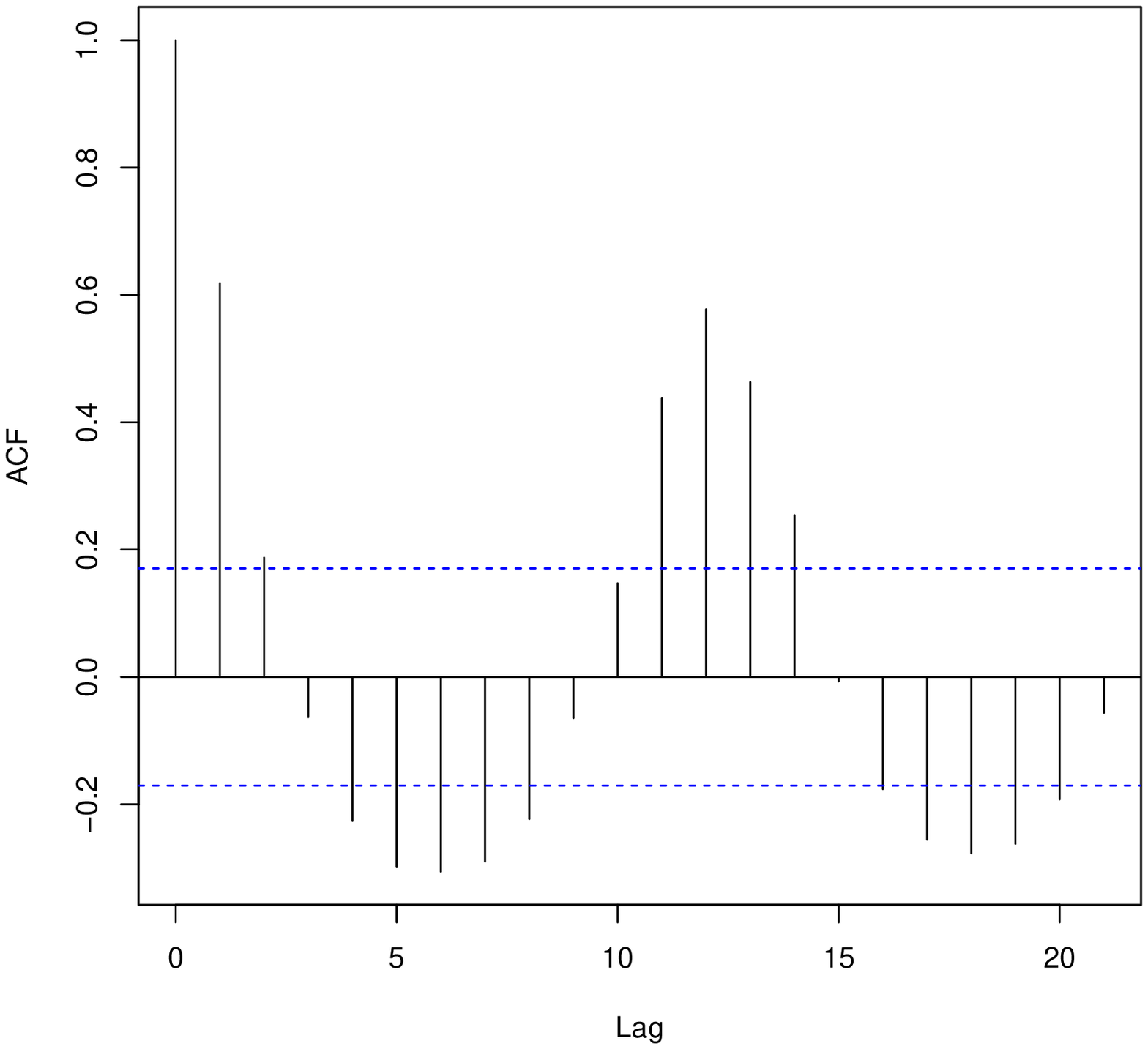}
\caption{Auto-correlation function of Monthly pneumonia and influenza deaths time series.}
\label{acf_flu.eps}
\end{figure}

Figure \ref{acf_flu.eps} of auto-correlation at the appendix shows a seasonality of approximately 12 months. In this line the predictor is based on a auto-regressive model of order $p = 12$, this is $r(z_k)= z_k = [y_{k-1} \;y_{k-2}\;...\;y_{k-12}]^T$.

\begin{figure}[!ht]
\centering
\includegraphics [width=80mm,height=70mm]{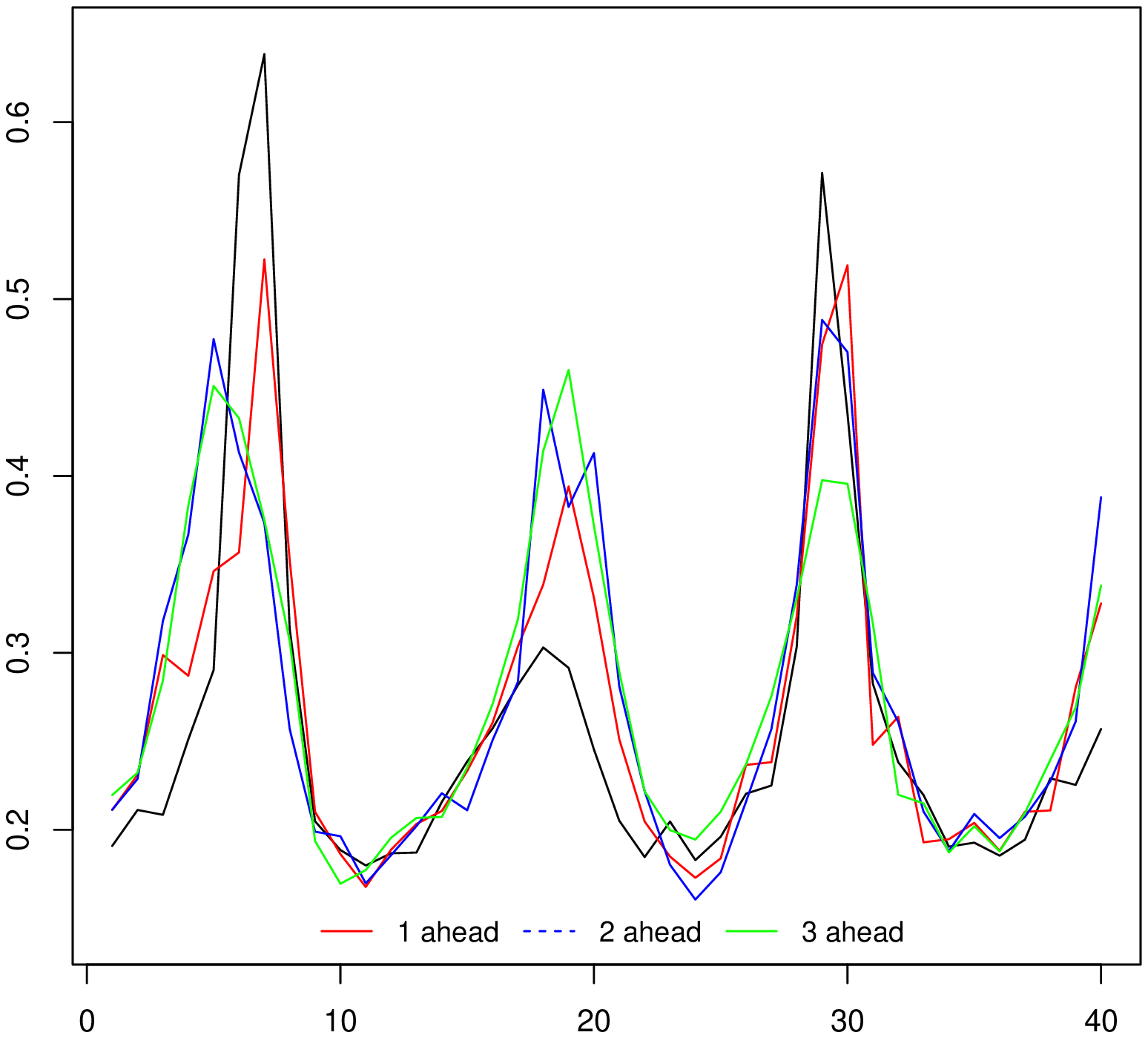}
\caption{Monthly pneumonia and influenza deaths time series predictions by forecasting horizon in the test set.}
\label{fig:flu_pred_NR}
\end{figure}

Figure \ref{fig:flu_pred_NR} shows the forecasts of the proposed predictor by forecasting horizons with the hyper-parameters selected in both error measures.

Results of this time series are shown on Table \ref{table:flu_table} at the appendix. The hyper-parameter $\gamma$ is selected in the training set where the error is minimum, the value of $\gamma$ is inferred to perform forecasts in the test set as shown in table \ref{table_gamma_flu}.

\begin{table}[!ht]
\caption{Monthly pneumonia and influenza deaths time series optimal gamma.}
\label{table_gamma_flu}
\centering
\begingroup\fontsize{8pt}{8pt}\selectfont
\begin{tabular}{lrrr}
  \hline
Ahead & $\gamma$\_mape & $\gamma$\_smape\\ 
  \hline
1.00 & 0.50 & 0.50 \\ 
2.00 & 0.22 & 0.22 \\ 
3.00 & 0.08 & 0.08 \\ 
   \hline
\end{tabular}\par
\endgroup
\end{table}

\begin{figure}[!ht]
\begin{subfigure}{0.5\textwidth}
\includegraphics[width=1.01\linewidth, height=50mm]{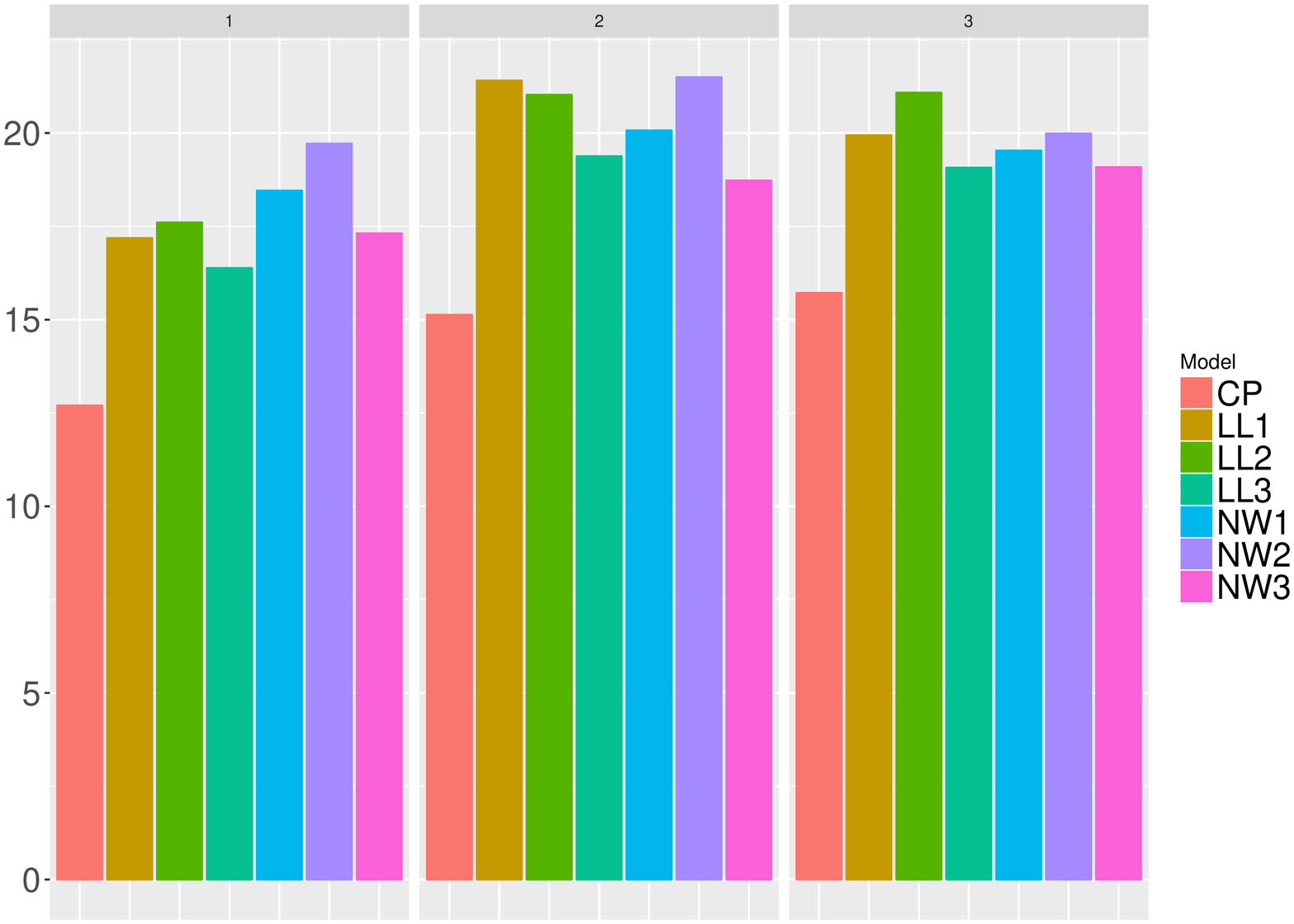}
\vspace*{-4mm}
\caption{SMAPE}
\label{flu_SMAPE.eps}
\end{subfigure}
\begin{subfigure}{0.5\textwidth}
\includegraphics[width=1.01\linewidth, height=50mm]{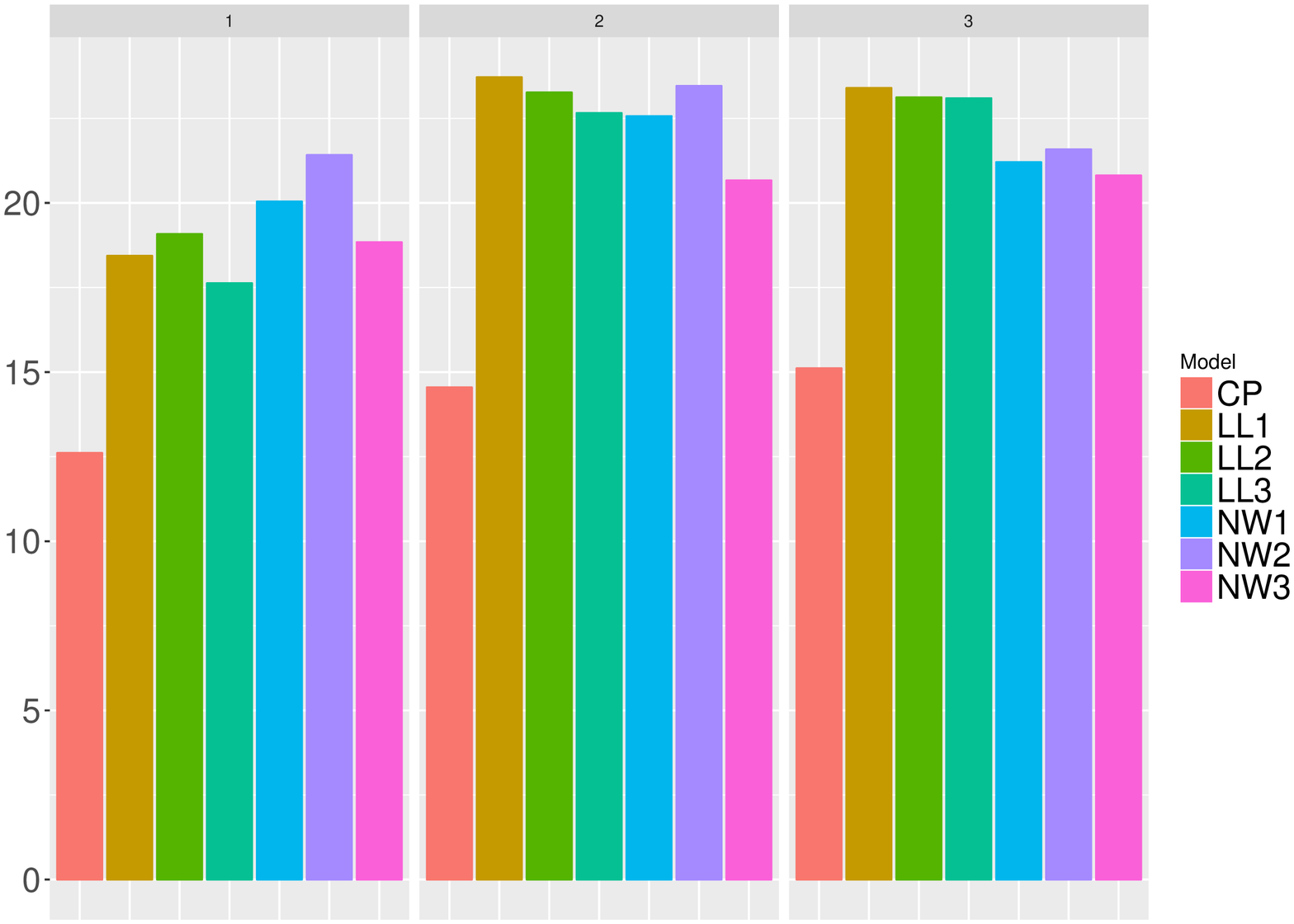}
\vspace*{-4mm}
\caption{MAPE}
\label{flu_MAPE.eps}
\end{subfigure}
\caption{Mean of test set errors by forecasting horizon in monthly pneumonia and influenza deaths time series.}
\label{fig:barplot_flu}
\end{figure}

To sum up table \ref{table:flu_table} in a graphical way, figure \ref{fig:barplot_flu} plots the error measures by predictor and prediction horizon.

Results shows by forecasting horizon that the proposed predictor outperforms in both error measures in the proposed forecast horizons. Besides, the results shows that tricube kernel is a suitable option that outperforms between the other selected non parametric methods for a short term forecast

\subsubsection{Averaged results}\label{avg_results}
This subsection averages the results shown in tables \ref{table:Air_table}, \ref{table:Lynx_table}, \ref{table:Rad_table} and \ref{table:flu_table} attached at the appendix. This results, averages all time series test error results in two error measures selection criterion, depending on the selected criteria in the training set the results could vary. Results shows that in average the proposed predictor outperforms the selected methods in the different results.

\begin{figure}[!ht]
\begin{subfigure}{0.5\textwidth}
\includegraphics[width=1.01\linewidth, height=50mm]{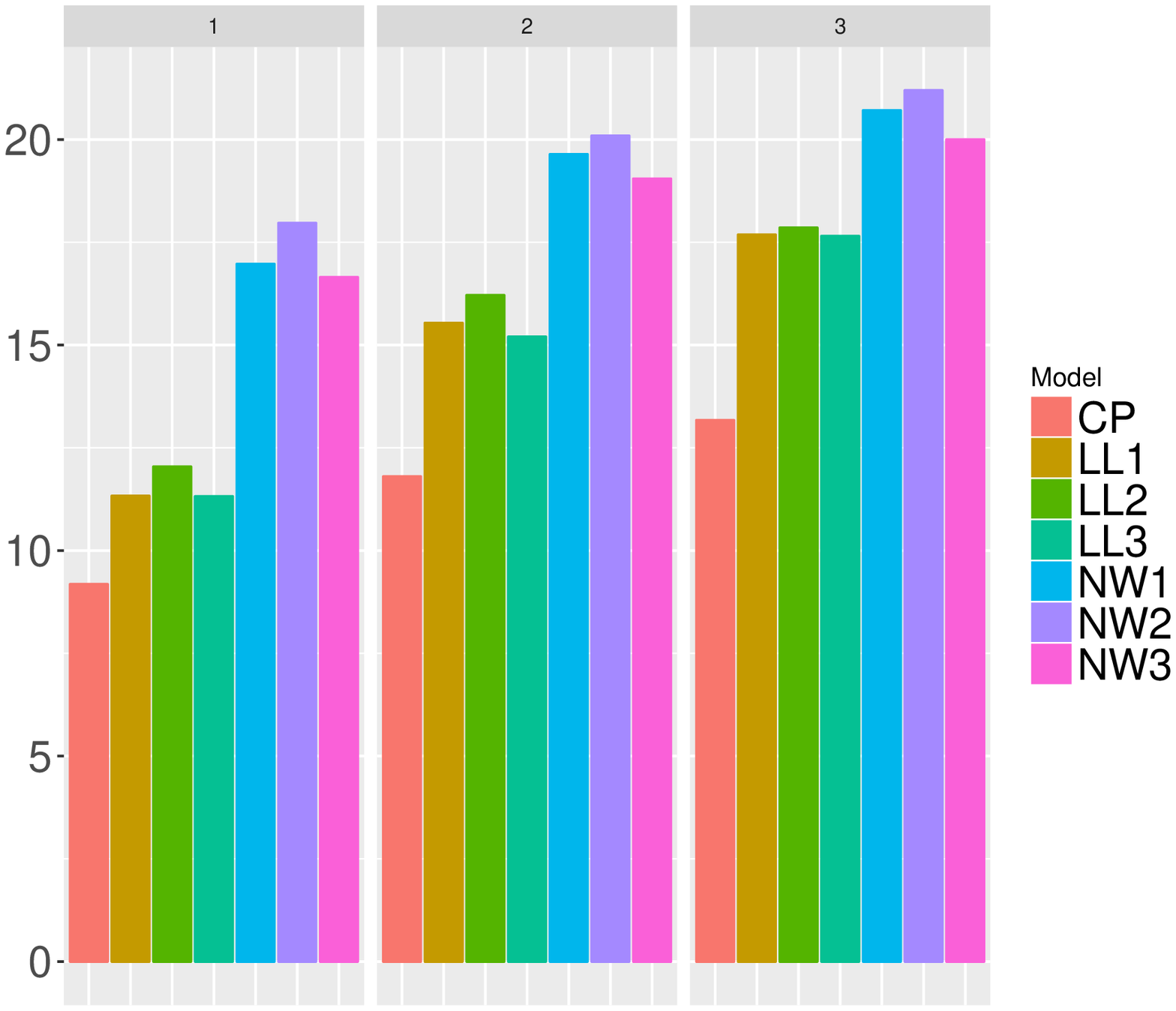}
\vspace*{-4mm}
\caption{SMAPE}
\label{NR_MAPE.eps}
\end{subfigure}
\begin{subfigure}{0.5\textwidth}
\includegraphics[width=1.01\linewidth, height=50mm]{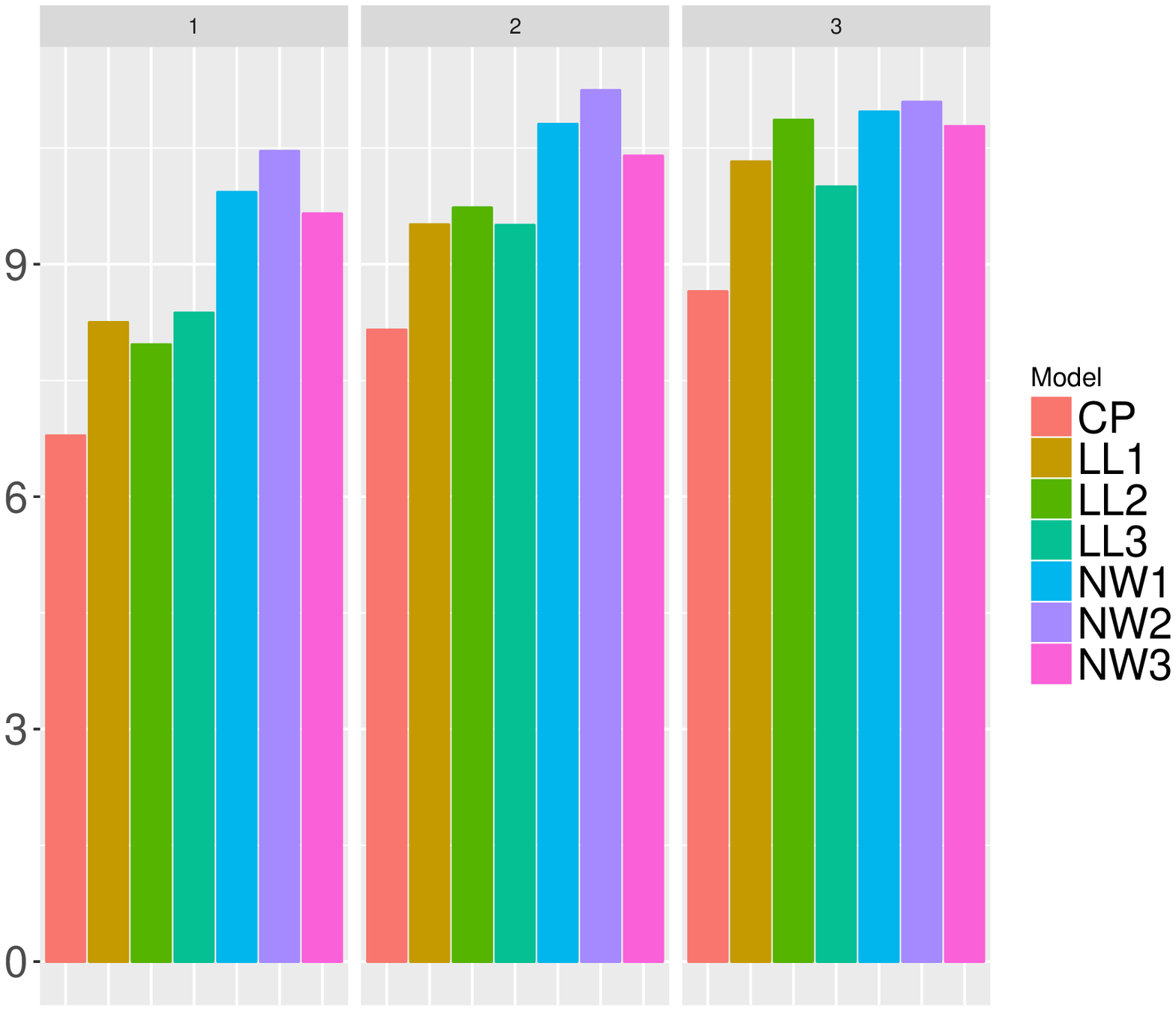}
\vspace*{-4mm}
\caption{MAPE}
\label{NR_SMAPE.eps}
\end{subfigure}
\caption{Mean of test set errors by forecasting horizon}
\label{fig:NR_AGG_PLOT}
\end{figure}

\begin{figure}
\centering
\includegraphics [width=100mm,height=70mm]{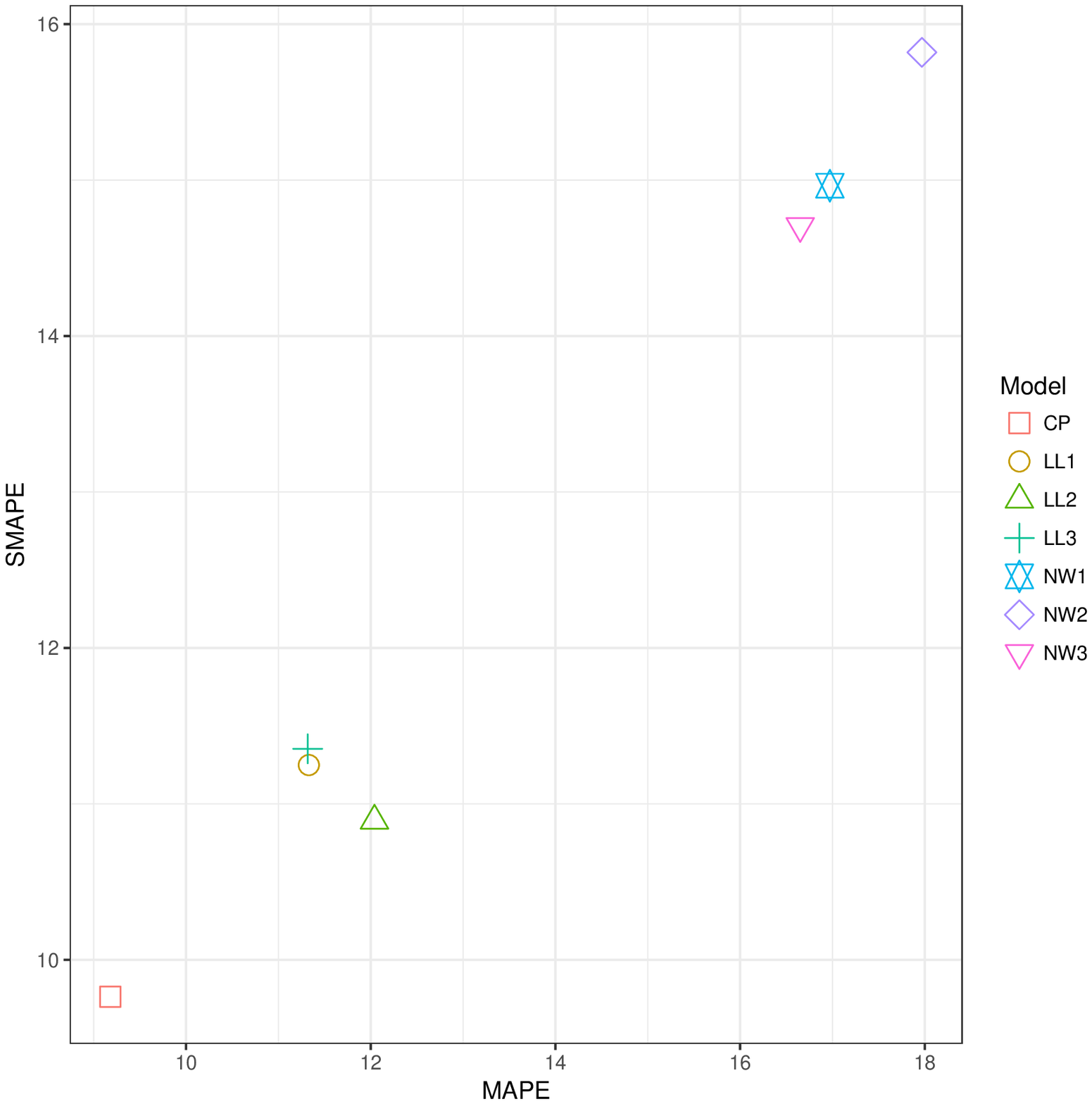}
\caption{Mean of SMAPE and MAPE results for 1 step-ahead forecasts}
\label{NR_1.eps}
\end{figure}

\begin{figure}
\centering
\includegraphics [width=100mm,height=70mm]{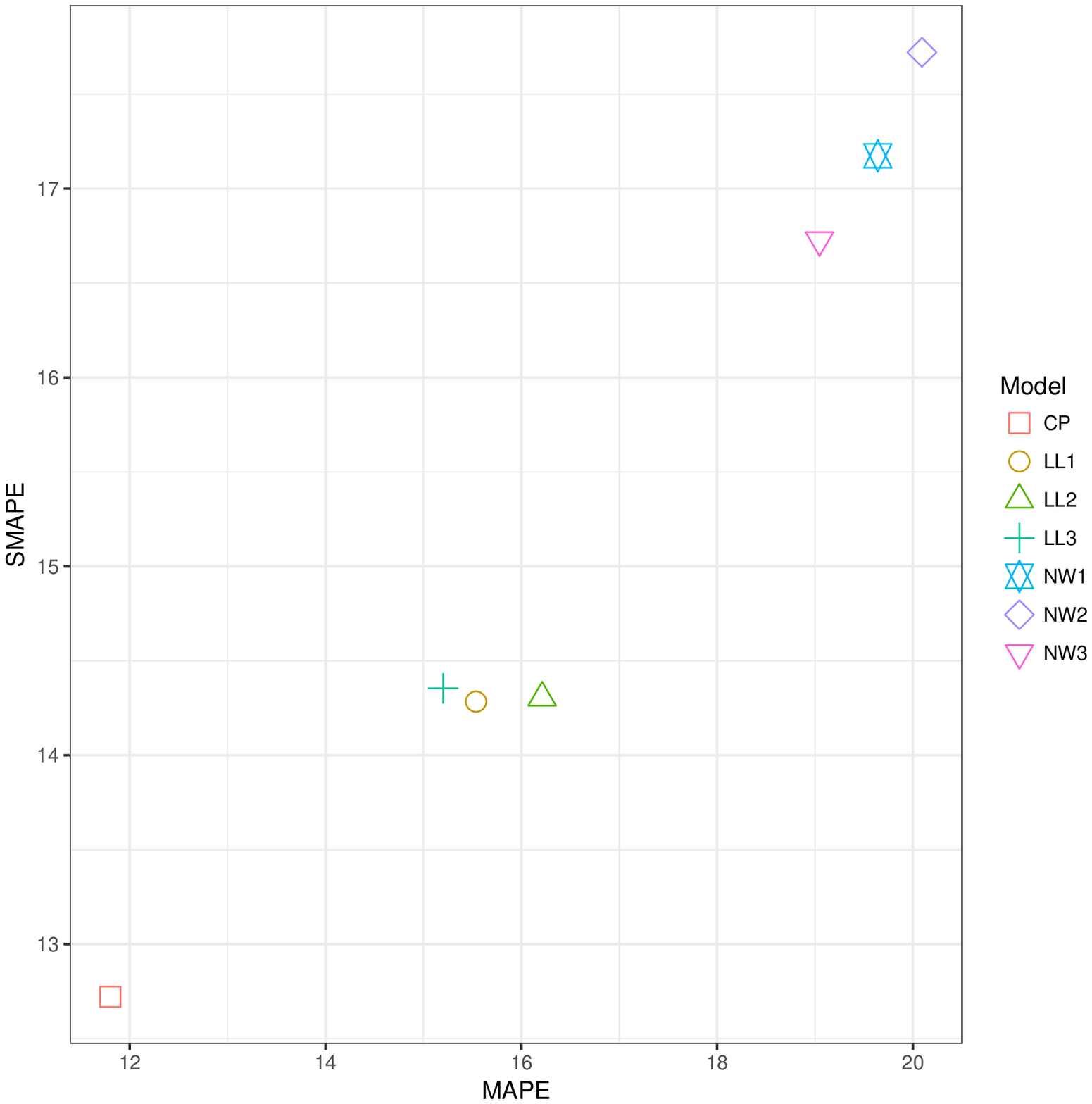}
\caption{Mean of SMAPE and MAPE results for 2 step-ahead forecasts}
\label{NR_2.eps}
\end{figure}

\begin{figure}
\centering
\includegraphics [width=100mm,height=70mm]{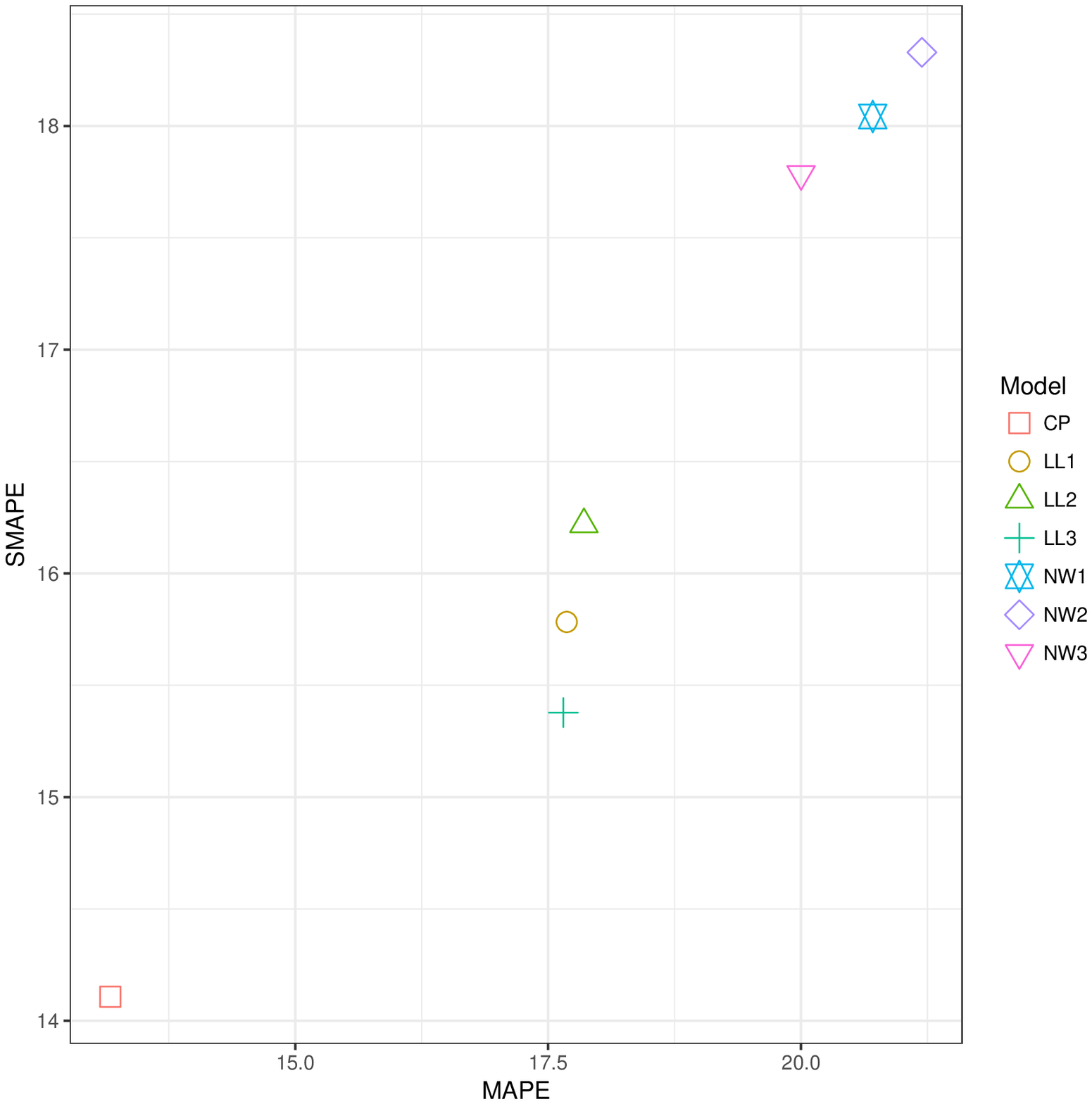}
\caption{Mean of SMAPE and MAPE results for 3 step-ahead forecasts}
\label{NR_3.eps}
\end{figure}

Figures from \ref{NR_1.eps} to \ref{NR_3.eps} with a different aggregation level show an average by grouping results in the different forecasting horizons shows that the proposed predictor is outperforming the proposed methods and offering a competitive alternative model.

\subsection{Monthly electricity supplied}\label{electricity}
It is provided in this subsection a real-
world dataset in order to perform the proposed predictor. This is, the IEA provides monthly statistics with timely and consistent oil, oil price, natural gas and electricity data for all Organization for Economic Co-operation and Development member countries.

Countries submitted monthly data is adjusted proportionately to maintain consistency with the most recent annual data for each generation source.

This time series is the electricity supplied for Spain from January of $2000$ to May of $2017$, this data consist in Indigenous production plus Imports minus Exports. It includes transmission and distribution losses.

\begin{figure}[!ht]
\begin{subfigure}{0.5\textwidth}
\includegraphics[width=1\linewidth, height=60mm]{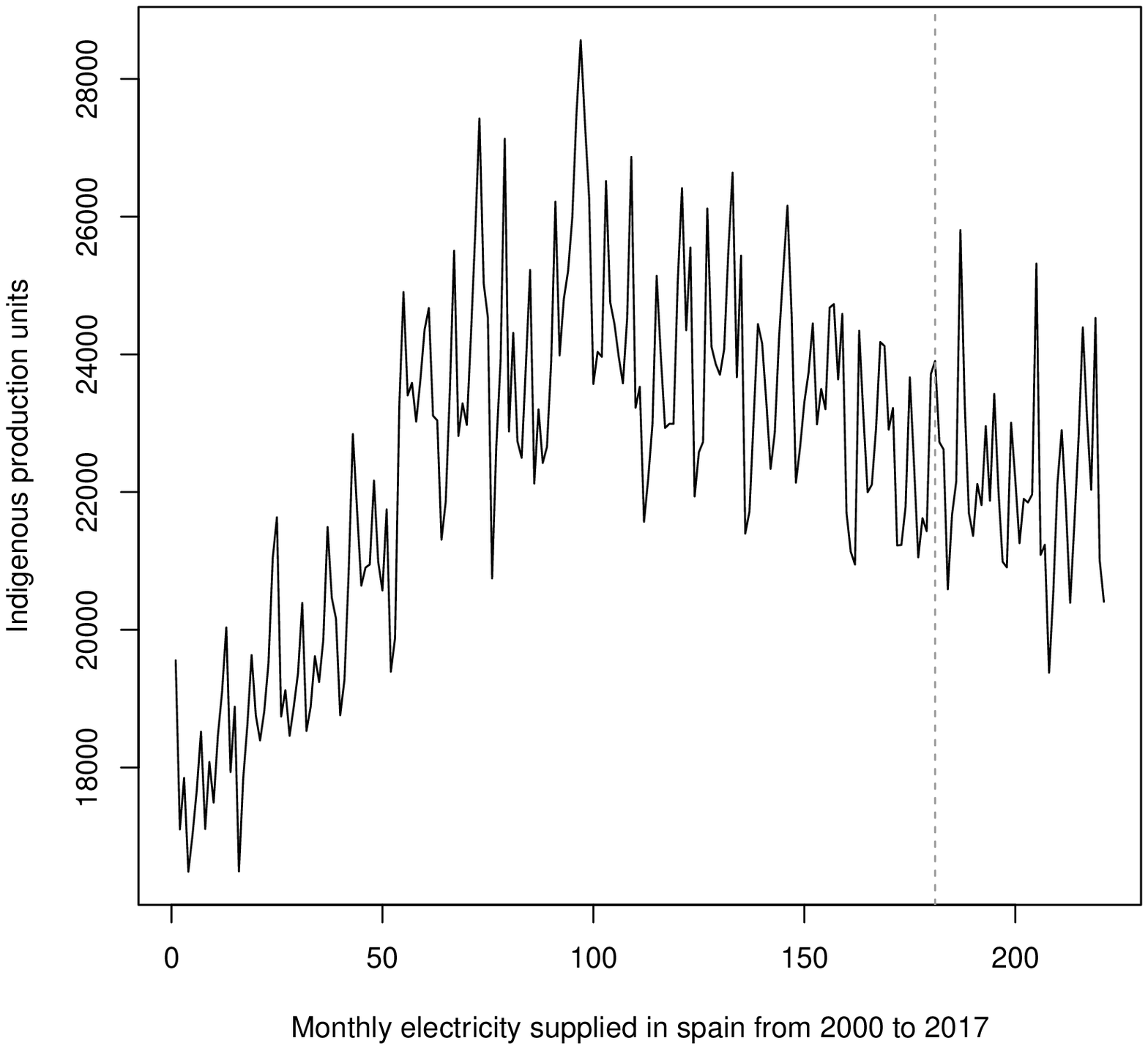}
\vspace*{-4mm}
\caption{Original Time series}
\label{elect.eps}
\end{subfigure}
\begin{subfigure}{0.5\textwidth}
\includegraphics[width=1\linewidth, height=60mm]{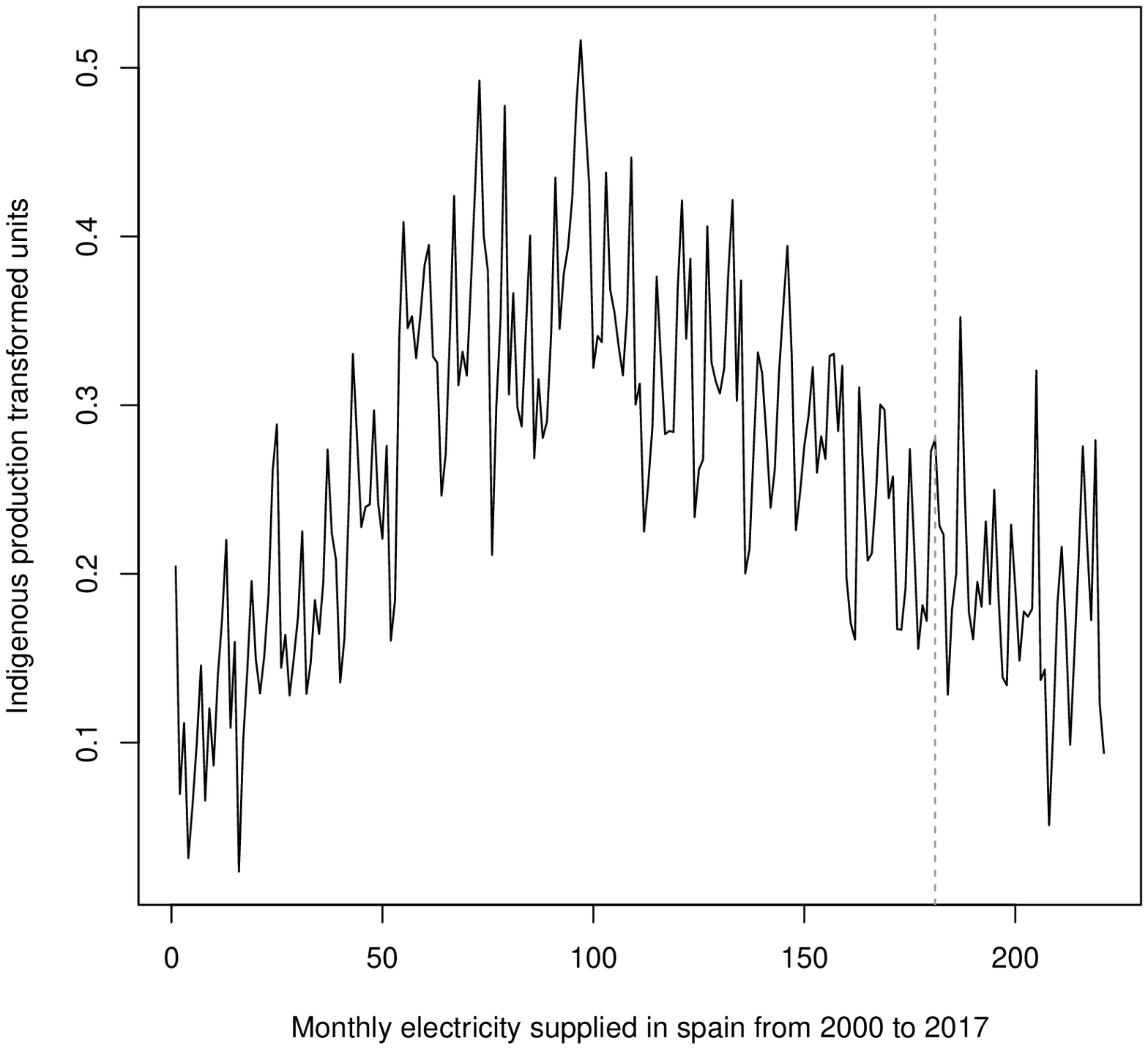}
\vspace*{-4mm}
\caption{Transformed Time series}
\label{elect_transformed.eps}
\end{subfigure}
\caption{Monthly electricity supplied in spain (2000$-$2017).}
\label{fig:elect_time}
\end{figure}

Figure \ref{fig:elect_time} plots the aforementioned data set which has $221$ observations, the first $181$ observations were used as training set and the last $40$ as a test set.

\begin{figure}[!ht]
\centering
\includegraphics [width=70mm,height=60mm]{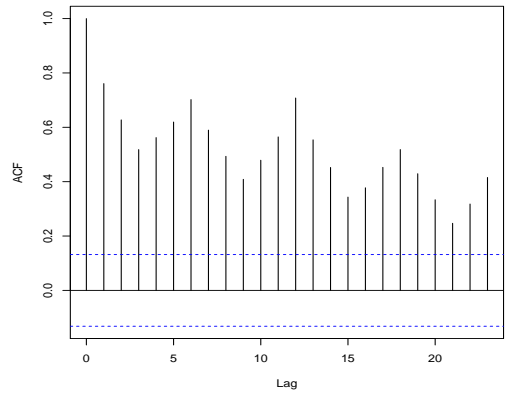}
\caption{Auto-correlation function of Monthly electricity supplied in spain.}
\label{acf_elect.eps}
\end{figure}

Figure \ref{acf_elect.eps} of auto-correlation at the appendix shows a seasonality of approximately 12 months. In this line the predictor is based on a auto-regressive model of order $p = 12$, this is $r(z_k)= z_k = [y_{k-1} \;y_{k-2}\;...\;y_{k-12}]^T$.

\begin{figure}[!ht]
\begin{subfigure}{0.5\textwidth}
\includegraphics[width=1.01\linewidth, height=60mm]{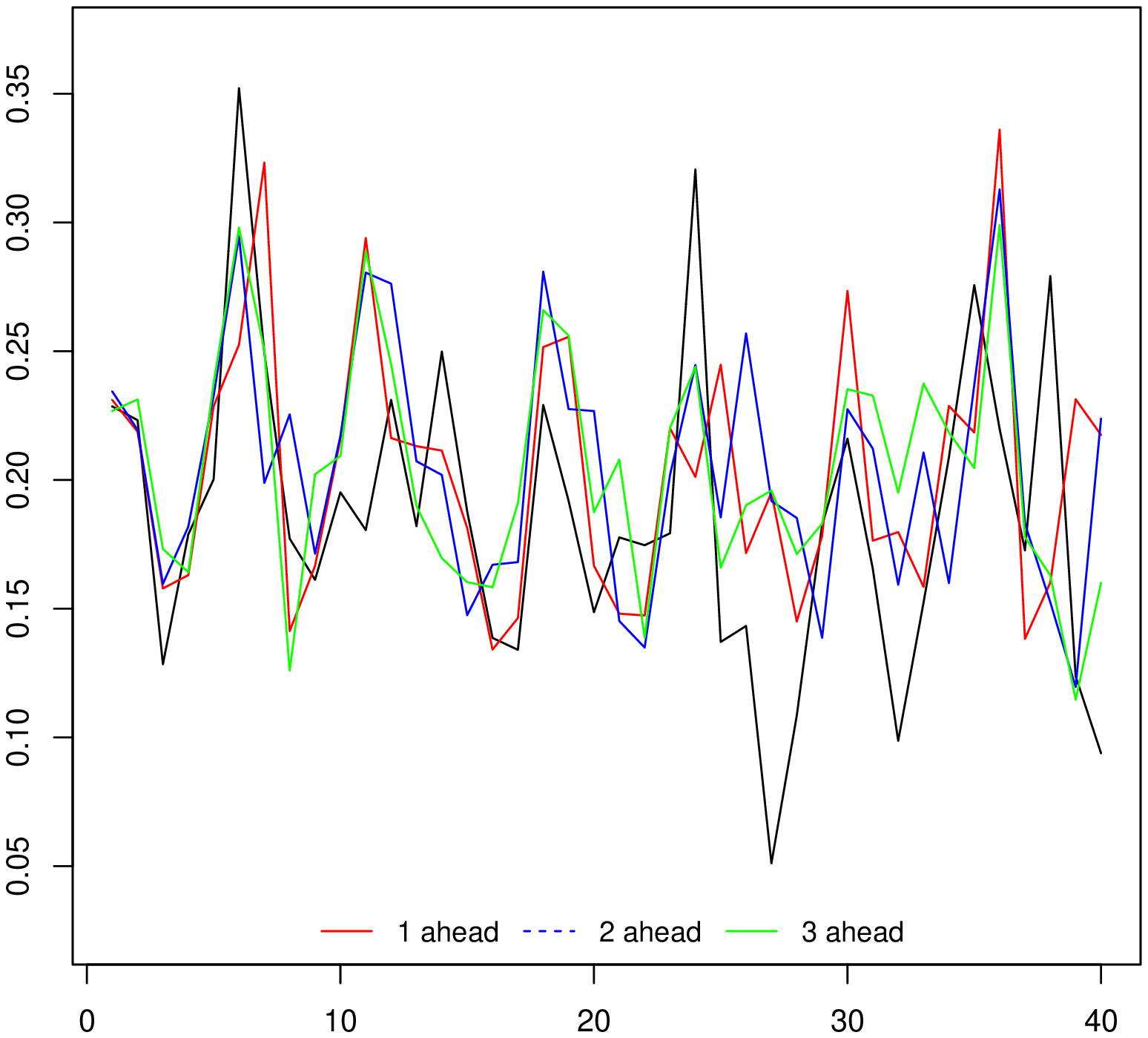}
\vspace*{-4mm}
\caption{MAPE selection criterion}
\label{elect_pred_MAPE.eps}
\end{subfigure}
\begin{subfigure}{0.5\textwidth}
\includegraphics[width=1.01\linewidth, height=60mm]{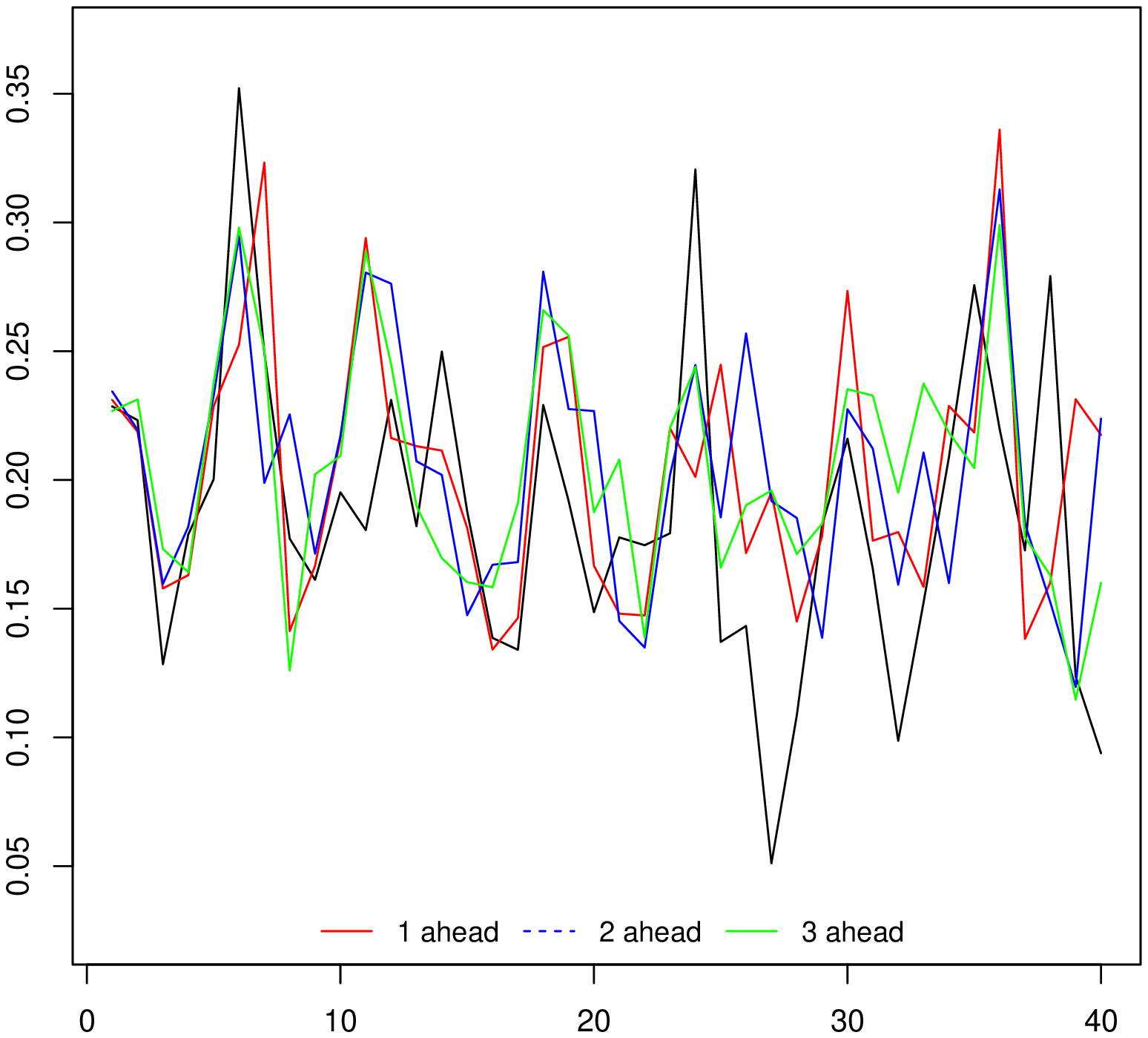}
\vspace*{-4mm}
\caption{SMAPE selection criterion}
\label{elect_pred_SMAPE.eps}
\end{subfigure}
\caption{Monthly electricity supplied time series predictions by forecasting horizon in the test set.}
\label{fig:elect_pred_NR}
\end{figure}

The forecasts for the proposed predictor are in figure \ref{fig:elect_pred_NR}, these are plotted by forecasting horizons with the hyper-parameters selected in both error measures.

Results of this time series are shown on Table \ref{table:elect_table} at the appendix. The hyper-parameter $\gamma$ is selected in the training set where the error is minimum, the value of $\gamma$ is inferred to perform forecasts in the test set as shown in table \ref{table_gamma_elect}.

\begin{table}[!htbp]
\caption{Monthly electricity supplied time series optimal gamma.}
\label{table_gamma_elect}
\centering
\begingroup\fontsize{8pt}{8pt}\selectfont
\begin{tabular}{lrrr}
  \hline
Ahead & $\gamma$\_mape & $\gamma$\_smape\\ 
  \hline
1.00 & 0.06 & 0.07 \\ 
2.00 & 0.26 & 0.25 \\ 
3.00 & 0.09 & 0.09 \\ 
   \hline
\end{tabular}\par 
\endgroup
\end{table}

\begin{figure}[!ht]
\begin{subfigure}{0.5\textwidth}
\includegraphics[width=1.01\linewidth, height=50mm]{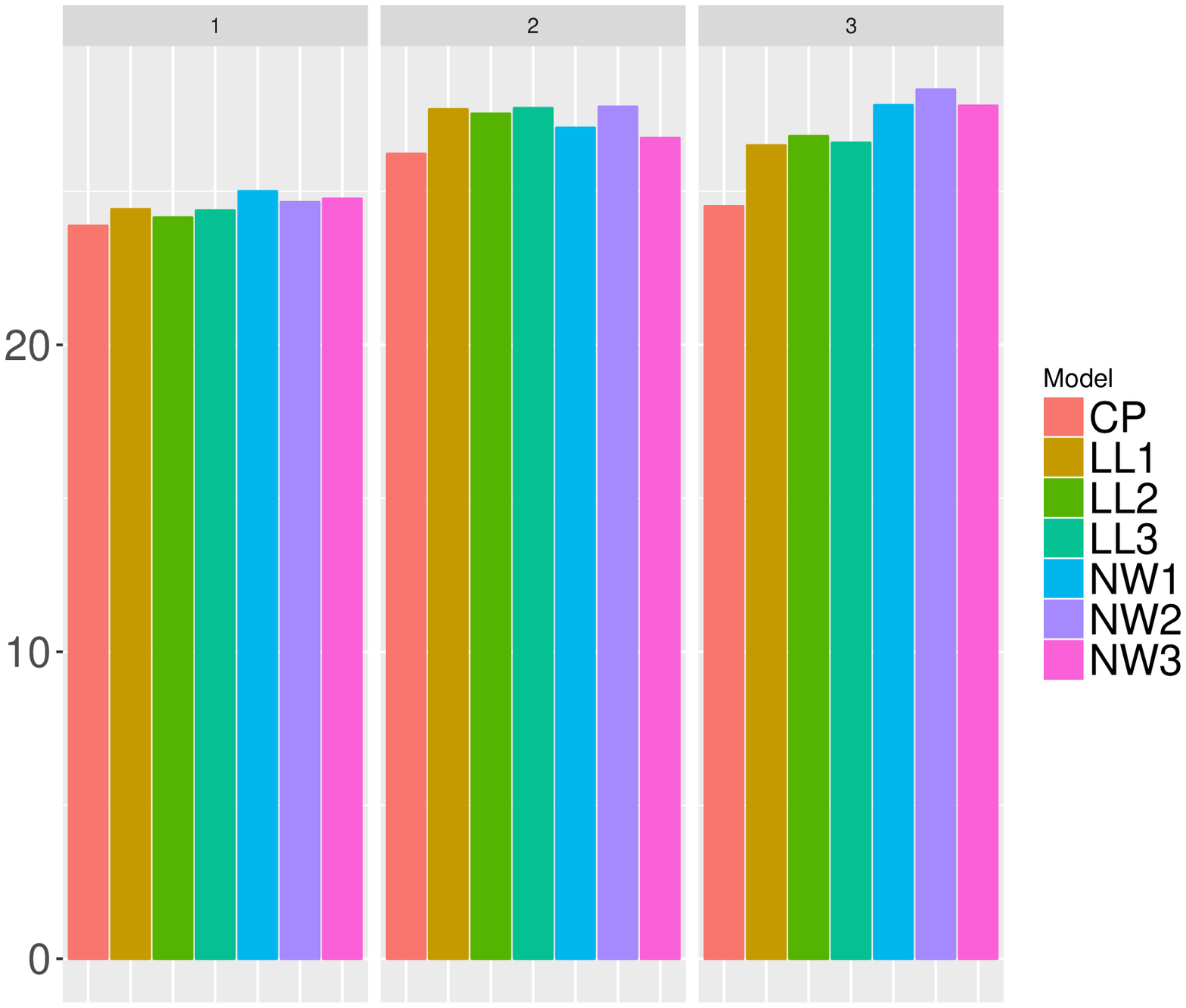}
\vspace*{-4mm}
\caption{SMAPE}
\label{elect_SMAPE.eps}
\end{subfigure}
\begin{subfigure}{0.5\textwidth}
\includegraphics[width=1.01\linewidth, height=50mm]{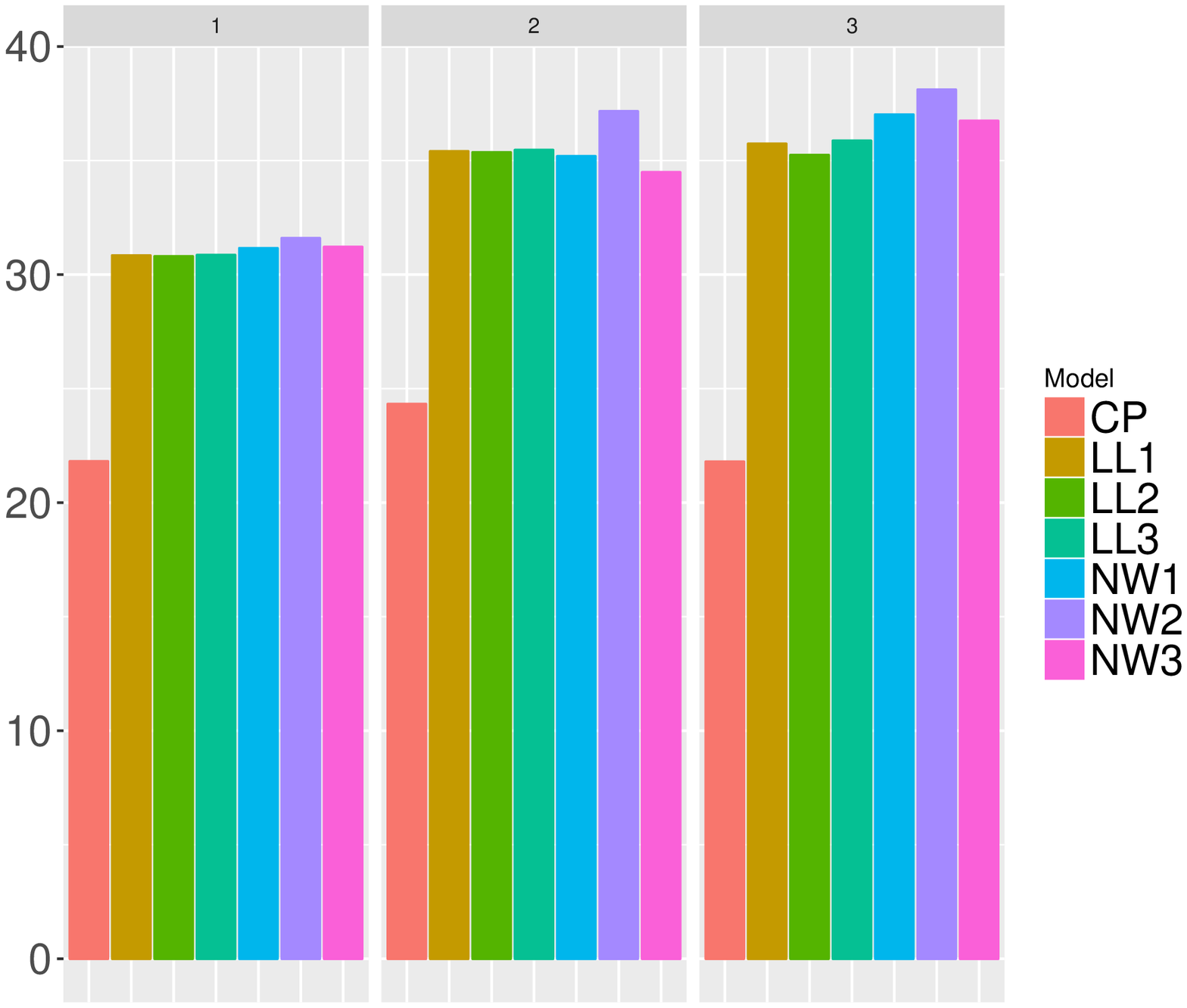}
\vspace*{-4mm}
\caption{MAPE}
\label{elect_MAPE.eps}
\end{subfigure}
\caption{Mean of test set errors by forecasting horizon in Monthly electricity supplied time series.}
\label{fig:barplot_elect}
\end{figure}

To sum up table \ref{table:elect_table} in a graphical way, figure \ref{fig:barplot_elect} plots the error measures by predictor and prediction horizon.

Results from Figure \ref{fig:barplot_elect} shows that the proposed predictor outperforms in both error measures in the proposed forecast horizons. The proposed predictor is a suitable option to consider in a real life problem.

\section{Conclusions}\label{conclusion}
A novel non-parametric Time Series forecasting method has been proposed. The prediction is obtained by a weighted sum of past observations. A combination of deterministic and stochastic assumptions are used to obtain an expression of the outer bound of the prediction error. The weights are obtained solving a convex optimization problem that minimizes the upper bound of the prediction error. The method includes a tuning hyper-parameter. This hyper-parameter may balance the deterministic and stochastic considered assumptions. By a cross-validation scheme, a suitable hyper-parameter can be obtained. The performance of the proposed predictor is exposed by some datasets. 

\clearpage
\section{Appendix}\label{appendix}
The following section contains the mathematical proofs, as well as research results contained in tables and some summary plots of this tables to make it easier for the reader to navigate through the document.

\subsection{Mathematical derivations}\label{Mathderiv}
Taking into account Assumption \ref{ASS_1} and Definitions \ref{def3} and \ref{def2} the following equalities can be inferred
\begin{eqnarray}
\hat{e}_k(\Psi) &=& y_{k+1}- \hat{y}_{k+1}(\Psi)     \\
&=& y_{k+1}- \Psi^T b_Y \\
&=&  r(z_k)^T \Phi_k - \Psi^T b_Y + e_k\\
&=&  (A^T\Psi)^T \Phi_k - \Psi^T b_Y + e_k \\
&=&  \Psi^T(A\Phi_k-b_Y) + e_k \\
&=&  \displaystyle\sum\limits_{j=1}^{k}\Psi_{j}(r(z_{j-1})^T\Phi_k-y_j ) + e_k \\
&=&  -\displaystyle\sum\limits_{j=1}^{k}\Psi_{j} e_{j-1} + e_k. \\
\end{eqnarray}
\emph{QED}

\subsection{Tables}\label{Tables}
\begin{table}
\centering
\begingroup\fontsize{8pt}{8pt}\selectfont
\begin{tabular}{llrrrrrrr}
  \hline
Model  & Ahead & $\gamma$ & tr\_MAPE & te\_MAPE & $\gamma$ & tr\_SMAPE & te\_SMAPE \\ 
  \hline
CP  & 1.00 & \textbf{0.12} & 15.75 & 13.04 & \textbf{0.12} & 15.46 & 14.97 \\ 
  LL1  & 1.00 & 1.57 & 15.19 & 14.65 & 1.84 & 14.34 & 15.69 \\ 
  LL2  & 1.00 & 1.55 & 14.82 & 17.11 & 1.62 & 13.77 & 14.13 \\ 
  LL3  & 1.00 & 1.89 & 16.42 & 15.47 & 2.33 & 14.04 & 16.98 \\ 
  NW1  & 1.00 & 1.08 & 17.69 & 27.26 & 1.08 & 13.82 & 21.05 \\ 
  NW2  & 1.00 & 1.02 & 17.36 & 28.38 & 1.02 & 13.48 & 21.90 \\ 
  NW3  & 1.00 & 1.24 & 17.91 & 27.16 & 1.26 & 13.98 & 21.09 \\ 
  CP  & 2.00 & \textbf{0.14} & 17.36 & 14.05 & \textbf{0.14} & 16.88 & 16.58 \\ 
  LL1  & 2.00 & 1.58 & 13.69 & 18.89 & 1.58 & 13.56 & 16.45 \\ 
  LL2  & 2.00 & 1.54 & 14.22 & 22.02 & 1.69 & 13.24 & 17.71 \\ 
  LL3  & 2.00 & 1.67 & 14.79 & 18.37 & 2.09 & 14.38 & 18.44 \\ 
  NW1  & 2.00 & 1.12 & 17.06 & 29.65 & 1.14 & 13.29 & 22.91 \\ 
  NW2  & 2.00 & 0.98 & 16.71 & 30.40 & 0.98 & 13.05 & 23.22 \\ 
  NW3  & 2.00 & 1.26 & 17.33 & 29.41 & 1.28 & 13.40 & 22.61 \\ 
  CP  & 3.00 & \textbf{0.00} & 17.43 & 15.45 & \textbf{0.00} & 16.50 & 18.66 \\ 
  LL1  & 3.00 & 1.64 & 16.06 & 24.99 & 2.15 & 15.59 & 21.13 \\ 
  LL2  & 3.00 & 1.61 & 17.89 & 25.89 & 2.01 & 15.37 & 22.14 \\ 
  LL3  & 3.00 & 1.78 & 16.49 & 25.33 & 2.34 & 15.72 & 20.72 \\ 
  NW1  & 3.00 & 1.12 & 17.16 & 32.13 & 1.14 & 13.23 & 24.05 \\ 
  NW2  & 3.00 & 0.98 & 16.88 & 32.80 & 0.98 & 13.13 & 24.09 \\ 
  NW3  & 3.00 & 1.24 & 17.55 & 31.12 & 1.30 & 13.37 & 23.75 \\ 
   \hline
\end{tabular}
\endgroup
\caption{Airline passengers time series results}
\label{table:Air_table}
\end{table}

\clearpage

\begin{table}
\centering
\begingroup\fontsize{8pt}{8pt}\selectfont
\begin{tabular}{llrrrrrrr}
  \hline
Model  & Ahead & $\gamma$ & tr\_MAPE & te\_MAPE & $\gamma$ & tr\_SMAPE & te\_SMAPE \\ 
  \hline
CP  & 1.00 & \textbf{0.01} & 6.75 & 5.09 & \textbf{0.06} & 6.73 & 5.13 \\ 
  LL1  & 1.00 & 13.01 & 6.66 & 4.94 & 13.01 & 6.62 & 5.03 \\ 
  LL2  & 1.00 & 12.01 & 6.54 & 5.00 & 12.01 & 6.50 & 5.09 \\ 
  LL3  & 1.00 & 16.01 & 6.72 & 4.86 & 16.01 & 6.68 & 4.94 \\ 
  NW1  & 1.00 & 4.33 & 8.42 & 10.53 & 4.47 & 7.99 & 10.98 \\ 
  NW2  & 1.00 & 4.03 & 8.28 & 10.40 & 4.03 & 7.87 & 10.88 \\ 
  NW3  & 1.00 & 5.11 & 8.42 & 10.69 & 5.27 & 7.99 & 11.18 \\ 
  CP  & 2.00 & \textbf{0.00} & 10.08 & 9.31 & \textbf{0.02} & 10.07 & 8.88 \\ 
  LL1  & 2.00 & 11.01 & 9.94 & 9.41 & 11.01 & 9.85 & 9.79 \\ 
  LL2  & 2.00 & 11.01 & 9.82 & 9.29 & 12.01 & 9.66 & 8.83 \\ 
  LL3  & 2.00 & 12.01 & 9.99 & 9.67 & 12.01 & 9.89 & 10.11 \\ 
  NW1  & 2.00 & 4.76 & 9.45 & 12.29 & 4.77 & 8.91 & 12.87 \\ 
  NW2  & 2.00 & 4.31 & 9.33 & 11.83 & 4.31 & 8.79 & 12.39 \\ 
  NW3  & 2.00 & 5.49 & 9.46 & 12.41 & 5.66 & 8.92 & 13.04 \\ 
  CP  & 3.00 & \textbf{0.02} & 11.18 & 12.49 & \textbf{0.04} & 11.27 & 11.73 \\ 
  LL1  & 3.00 & 12.01 & 11.24 & 11.29 & 12.01 & 10.92 & 11.84 \\ 
  LL2  & 3.00 & 10.01 & 10.98 & 11.32 & 12.01 & 10.72 & 11.40 \\ 
  LL3  & 3.00 & 14.01 & 11.36 & 11.20 & 16.01 & 11.00 & 11.58 \\ 
  NW1  & 3.00 & 4.96 & 9.88 & 13.04 & 4.96 & 9.30 & 13.71 \\ 
  NW2  & 3.00 & 4.35 & 9.58 & 12.76 & 4.35 & 9.03 & 13.43 \\ 
  NW3  & 3.00 & 5.70 & 9.92 & 13.07 & 5.70 & 9.34 & 13.75 \\ 
   \hline
\end{tabular}
\endgroup
\caption{Canadian Lynx time series results}
\label{table:Lynx_table}
\end{table}

\clearpage

\begin{table}
\centering
\begingroup\fontsize{8pt}{8pt}\selectfont
\begin{tabular}{llrrrrrrr}
  \hline
Model  & Ahead & $\gamma$ & tr\_MAPE & te\_MAPE & $\gamma$ & tr\_SMAPE & te\_SMAPE \\ 
   \hline
CP  & 1.00 & \textbf{0.00} & 7.33 & 6.70 & \textbf{0.00} & 7.26 & 6.89 \\ 
  LL1  & 1.00 & 50.00 & 7.25 & 7.29 & 50.00 & 7.23 & 7.09 \\ 
  LL2  & 1.00 & 30.00 & 7.17 & 6.98 & 30.00 & 7.18 & 6.74 \\ 
  LL3  & 1.00 & 60.00 & 7.26 & 7.31 & 60.00 & 7.24 & 7.10 \\ 
  NW1  & 1.00 & 13.00 & 8.92 & 10.06 & 13.00 & 8.92 & 9.36 \\ 
  NW2  & 1.00 & 13.00 & 9.09 & 11.68 & 13.00 & 9.04 & 10.79 \\ 
  NW3  & 1.00 & 15.00 & 8.86 & 9.91 & 15.00 & 8.85 & 9.24 \\ 
  CP  & 2.00 & \textbf{0.03} & 11.16 & 8.71 & \textbf{0.00} & 10.92 & 9.58 \\ 
  LL1  & 2.00 & 40.00 & 10.75 & 10.13 & 40.00 & 10.74 & 9.50 \\ 
  LL2  & 2.00 & 40.00 & 10.71 & 10.27 & 40.00 & 10.69 & 9.66 \\ 
  LL3  & 2.00 & 50.00 & 10.74 & 10.11 & 50.00 & 10.74 & 9.49 \\ 
  NW1  & 2.00 & 14.00 & 10.51 & 14.07 & 14.00 & 10.45 & 12.84 \\ 
  NW2  & 2.00 & 12.00 & 10.60 & 14.68 & 13.00 & 10.52 & 13.78 \\ 
  NW3  & 2.00 & 16.00 & 10.49 & 13.69 & 16.00 & 10.44 & 12.53 \\ 
  CP  & 3.00 & \textbf{0.00} & 12.42 & 9.63 & \textbf{0.00} & 12.25 & 10.34 \\ 
  LL1  & 3.00 & 50.00 & 12.26 & 11.05 & 50.00 & 12.14 & 10.22 \\ 
  LL2  & 3.00 & 40.00 & 12.18 & 11.08 & 40.00 & 12.07 & 10.25 \\ 
  LL3  & 3.00 & 50.00 & 12.24 & 10.98 & 50.00 & 12.14 & 10.15 \\ 
  NW1  & 3.00 & 14.00 & 11.65 & 16.46 & 14.00 & 11.53 & 14.87 \\ 
  NW2  & 3.00 & 13.00 & 11.74 & 17.64 & 13.00 & 11.57 & 15.81 \\ 
  NW3  & 3.00 & 15.00 & 11.61 & 15.00 & 16.00 & 11.52 & 14.56 \\ 
   \hline
\end{tabular}
\endgroup
\caption{Monthly critical radio frequencies time series results}
\label{table:Rad_table}
\end{table}

\begin{table}
\centering
\begingroup\fontsize{8pt}{8pt}\selectfont
\begin{tabular}{llrrrrrrr}
  \hline
Model  & Ahead & $\gamma$ & tr\_MAPE & te\_MAPE & $\gamma$ & tr\_SMAPE & te\_SMAPE \\ 
   \hline
CP  & 1.00 & \textbf{0.50} & 10.30 & 11.89 & \textbf{0.50} & 9.70 & 12.06 \\ 
  LL1  & 1.00 & 1.47 & 10.47 & 18.44 & 1.48 & 10.31 & 17.19 \\ 
  LL2  & 1.00 & 1.41 & 10.69 & 19.08 & 1.41 & 10.56 & 17.60 \\ 
  LL3  & 1.00 & 1.61 & 10.80 & 17.63 & 1.61 & 10.68 & 16.38 \\ 
  NW1  & 1.00 & 0.92 & 10.87 & 20.04 & 0.92 & 10.96 & 18.46 \\ 
  NW2  & 1.00 & 0.92 & 11.38 & 21.41 & 0.92 & 11.45 & 19.72 \\ 
  NW3  & 1.00 & 0.92 & 10.58 & 18.84 & 0.92 & 10.69 & 17.31 \\ 
  CP  & 2.00 & \textbf{0.22} & 12.81 & 15.14 & \textbf{0.22} & 12.04 & 15.85 \\ 
  LL1  & 2.00 & 1.95 & 13.99 & 23.72 & 1.48 & 13.95 & 21.40 \\ 
  LL2  & 2.00 & 1.90 & 13.54 & 23.27 & 1.90 & 13.77 & 21.02 \\ 
  LL3  & 2.00 & 1.61 & 14.70 & 22.66 & 1.55 & 14.08 & 19.38 \\ 
  NW1  & 2.00 & 1.01 & 11.22 & 22.56 & 0.96 & 11.24 & 20.07 \\ 
  NW2  & 2.00 & 0.95 & 11.65 & 23.46 & 0.95 & 11.67 & 21.50 \\ 
  NW3  & 2.00 & 0.92 & 10.84 & 20.67 & 0.92 & 10.78 & 18.73 \\ 
  CP  & 3.00 & \textbf{0.08} & 13.62 & 15.11 & \textbf{0.08} & 12.97 & 15.71 \\ 
  LL1  & 3.00 & 1.91 & 13.52 & 23.40 & 1.49 & 13.38 & 19.94 \\ 
  LL2  & 3.00 & 1.88 & 13.09 & 23.12 & 1.88 & 13.11 & 21.08 \\ 
  LL3  & 3.00 & 2.00 & 13.88 & 23.09 & 1.63 & 13.43 & 19.07 \\ 
  NW1  & 3.00 & 0.92 & 11.08 & 21.20 & 0.92 & 11.17 & 19.53 \\ 
  NW2  & 3.00 & 0.92 & 11.65 & 21.58 & 0.92 & 11.84 & 19.99 \\ 
  NW3  & 3.00 & 0.94 & 10.92 & 20.81 & 0.94 & 10.96 & 19.09 \\ 
   \hline
\end{tabular}
\endgroup
\caption{Monthly pneumonia and influenza deaths time series results}
\label{table:flu_table}
\end{table}

\begin{table}
\centering
\begingroup\fontsize{8pt}{8pt}\selectfont
\begin{tabular}{llrrrrrrr}
  \hline
Model  & Ahead & $\gamma$ & tr\_MAPE & te\_MAPE & $\gamma$ & tr\_SMAPE & te\_SMAPE \\ 
   \hline
CP &  1.00 & \textbf{0.06} & 13.88 & 21.81 & \textbf{0.07} & 14.16 & 23.88 \\ 
  LL1 &  1.00 & 1.90 & 17.08 & 30.84 & 2.00 & 14.52 & 24.42 \\ 
  LL2 &  1.00 & 1.87 & 16.93 & 30.80 & 1.50 & 14.40 & 24.14 \\ 
  LL3 &  1.00 & 2.00 & 17.14 & 30.86 & 2.00 & 14.62 & 24.38 \\ 
  NW1 &  1.00 & 0.68 & 17.76 & 31.16 & 0.73 & 15.00 & 25.00 \\ 
  NW2 &  1.00 & 0.65 & 17.58 & 31.60 & 0.65 & 15.04 & 24.65 \\ 
  NW3 &  1.00 & 0.78 & 17.81 & 31.21 & 0.81 & 14.98 & 24.76 \\ 
  CP &  2.00 & \textbf{0.26} & 15.22 & 24.32 & \textbf{0.25} & 15.23 & 26.22 \\ 
  LL1 &  2.00 & 2.00 & 18.01 & 35.40 & 2.00 & 15.94 & 27.67 \\ 
  LL2 &  2.00 & 1.98 & 17.77 & 35.36 & 1.85 & 15.85 & 27.53 \\ 
  LL3 &  2.00 & 2.00 & 18.18 & 35.46 & 2.00 & 16.09 & 27.71 \\ 
  NW1 &  2.00 & 0.70 & 19.01 & 35.19 & 0.72 & 16.14 & 27.07 \\ 
  NW2 &  2.00 & 0.68 & 18.96 & 37.16 & 0.70 & 16.24 & 27.75 \\ 
  NW3 &  2.00 & 0.78 & 19.04 & 34.48 & 0.80 & 16.19 & 26.74 \\ 
  CP &  3.00 & \textbf{0.09} & 15.95 & 21.80 & \textbf{0.09} & 16.13 & 24.51 \\ 
  LL1 &  3.00 & 1.96 & 18.62 & 35.73 & 1.92 & 16.30 & 26.50 \\ 
  LL2 &  3.00 & 1.79 & 18.33 & 35.24 & 1.59 & 16.02 & 26.80 \\ 
  LL3 &  3.00 & 2.00 & 18.76 & 35.87 & 2.00 & 16.48 & 26.58 \\ 
  NW1 & 3.00 & 0.73 & 19.92 & 37.01 & 0.73 & 16.87 & 27.81 \\ 
  NW2 & 3.00 & 0.65 & 19.24 & 38.11 & 0.65 & 16.70 & 28.31 \\ 
  NW3 & 3.00 & 0.83 & 20.01 & 36.74 & 0.85 & 16.92 & 27.79 \\ 
   \hline
\end{tabular}
\endgroup
\caption{Monthly electricity supplied in spain time series results}
\label{table:elect_table}
\end{table}

\clearpage

\bibliographystyle{oryx-apa}
\bibliography{refs}

\end{document}